\pdfoutput=1

\documentclass[11pt]{article}
\usepackage{acl}

\usepackage{microtype}
\usepackage{graphicx}
\usepackage{subfigure}
\usepackage{booktabs} 
\usepackage{hyperref}
\usepackage{algorithm}
\usepackage{algpseudocode}

\usepackage{colortbl}

\usepackage{amsmath}
\usepackage{amssymb}
\usepackage{mathtools}
\usepackage{amsthm}

\usepackage[capitalize,noabbrev]{cleveref}
\usepackage{colortbl}
\usepackage{nicematrix}
\usepackage[inline, shortlabels]{enumitem}
\usepackage{multirow}

\usepackage{amsthm}
\usepackage{amsmath,amsfonts,bm}
\usepackage{nicefrac}
\usepackage{array}
\usepackage{wrapfig}

\usepackage{hyperref}
\usepackage{url}
\usepackage{pbox}

\definecolor{bluex}{rgb}{0.27, 0.42, 0.81}
\definecolor{purplex}{HTML}{9564bf}
\definecolor{red3}{HTML}{C52A20}
\definecolor{red2}{HTML}{B36A6F}
\definecolor{red1}{HTML}{FFb5b5}
\definecolor{purple}{HTML}{B36A6F}
\definecolor{darkyellow}{HTML}{D5BA82}
\definecolor{blue1}{HTML}{508AB2}
\definecolor{blue2}{HTML}{C4E4E3}
\definecolor{green1}{HTML}{A1D0C7}
\definecolor{green2}{HTML}{BFF6BA}
\definecolor{green3}{HTML}{028100}
\definecolor{teal}{HTML}{508AB2}
\definecolor{Gray}{gray}{0.94}
\definecolor{orange3}{HTML}{c28c69}
\definecolor{blue3}{HTML}{3b75af}

\usepackage{pifont}
\newcommand{\cmark}{\text{\ding{51}}}
\newcommand{\xmark}{\text{\ding{55}}}

\usepackage[listings]{tcolorbox}
\tcbuselibrary{listings,theorems}
\newtcolorbox{mybox}{colback=white!5!white,colframe=black!75!black, left=.05in, right=.05in}

\newtcbtheorem{emptybox}{}%
{colback=green2!5,colframe=blue1,fonttitle=\bfseries,theorem style=plain, left=.0in, right=.0in,bottom=.02in, top=.02in,terminator sign none}{emptybox}
\newtcbtheorem[]{emptyboxx}{A Simple Template For Creating Backward Question}%
{colback=green2!5,colframe=blue1,fonttitle=\bfseries,theorem style=plain, left=.0in, right=.0in,bottom=.02in, top=.02in,terminator sign none}{emptyboxx}
\newtcbtheorem{bth}{Theoreme}
{colback=red!20,colframe=red,theorem style=plain,fonttitle=\bfseries,coltitle=black,terminator sign none}{th}

\newtcbtheorem[number within=section]{exmp}{Example}%
{colback=green2!5,colframe=blue1,fonttitle=\bfseries, left=.02in, right=.02in,bottom=.02in, top=.02in}{exmp}
\newtcbtheorem[]{prompt}{{\small Template For Creating Backward Question}}%
{colback=green!5,colframe=green!35!black,left=.0in, right=.0in,bottom=.02in, top=.02in,fonttitle=\bfseries}{th}
\newtcbtheorem[number within=section]{thm}{Theorem}%
{colback=green!5,colframe=green!35!black,fonttitle=\bfseries}{th}
\newtcbtheorem[number within=section]{corr}{Corollary}%
{colback=green!5,colframe=green!35!black,fonttitle=\bfseries}{th}
\newtcbtheorem{method}{Method}%
{colback=green!5,colframe=green!35!black,fonttitle=\bfseries}{th}

\newtcbtheorem{ebox}{}
{colback=green2!5,colframe=blue1, left=.02in, right=.02in,bottom=.02in,title empty, top=.02in, theorem style=plain apart,frame empty}{ebox}
\theoremstyle{plain}

\theoremstyle{definition}

\theoremstyle{remark}

\DeclareMathOperator*{\argmax}{arg\,max}

\newcommand{\bP}{\mathbb{P}}

\newcommand{\STAB}[1]{\begin{tabular}{@{}c@{}}#1\end{tabular}}

\newcommand*{\escape}[1]{\texttt{\textbackslash#1}}

\newcommand{\hA}{\mathcal{A}}

\newcommand{\hT}{\mathcal{T}}

\newcommand{\bI}{\mathbb{I}}

\newcommand{\vx}{{\bf x}}

\newcommand{\vP}{{\bf P}}

\newcommand*\demobox[1]{\pbox{.7\textwidth}{#1}}

\newcommand\Mark[1]{\textsuperscript#1}

\usepackage{latexsym}

\usepackage{times}
\usepackage{tablefootnote}
\usepackage{microtype}
\usepackage{inconsolata}
\usepackage[utf8]{inputenc}
\usepackage[
font = small,
labelfont = bf,
tableposition = top
]{caption}
\hypersetup{linkcolor=[rgb]{0.7,0.1,0.1}}
\hypersetup{citecolor=[HTML]{1663A9}}
\hypersetup{urlcolor=[HTML]{1663A9}}

\title{Forward-Backward Reasoning in Large Language Models \\
for Mathematical Verification}

\author{Weisen Jiang\Mark{{1, 2}}, Han Shi\Mark{3}, Longhui Yu\Mark{4}, Zhengying Liu\Mark{3} \\
	\bf Yu Zhang\Mark{{1, \thanks{Correspondence to: Yu Zhang}}}, Zhenguo Li\Mark{3}, James T. Kwok\Mark{2} \\
	\Mark{1}{\small{Department of Computer Science and Engineering, Southern University of Science and Technology}}  \\
	\Mark{2}{\small Department of Computer Science and Engineering, Hong Kong University of Science and Technology}\\
	\Mark{3}{\small Huawei Noah’s Ark Lab}
	\Mark{4}{\small Peking University}
	\\ 
	\texttt{\small \{waysonkong, yu.zhang.ust\}@gmail.com, jamesk@cse.ust.hk} \\ 
 {\small \textbf{Project page: \url{https://llm-fobar.github.io}}}
}

\begin{document}
\maketitle

\begin{abstract}
    Self-Consistency samples diverse reasoning chains with answers and chooses the final answer by majority voting. 
    It is based on forward reasoning and cannot further improve performance by sampling more reasoning chains when saturated.
    To further boost performance,
    we introduce backward reasoning 
    to verify candidate answers.
    Specifically, for mathematical tasks,
    we mask a number in the question and ask the LLM to answer a backward question created by a simple template, i.e., to predict the masked number when a candidate answer is provided.
    Instead of using forward or backward reasoning alone,
    we propose \textbf{FOBAR} to combine \textbf{FO}rward and \textbf{BA}ckward \textbf{R}easoning for verification.
    Extensive experiments on six standard mathematical data sets and three LLMs 
    show that FOBAR
    achieves state-of-the-art performance.
    In particular,
    FOBAR outperforms
    Self-Consistency, which
    uses forward reasoning alone,
    demonstrating that combining forward and backward reasoning is more accurate in verification.
    In addition,
    FOBAR achieves higher accuracy than existing verification methods,
    showing the effectiveness of the simple template used in backward reasoning and the proposed combination. 
\end{abstract}

\section{Introduction}
\label{sec:intro}
	
Pre-trained Large Language Models (LLMs) \citep{chowdhery2022palm, gpt4, wu2023openicl, jiang2023effective}
generalize well on unseen tasks by \textit{few-shot prompting} (or \textit{in-context learning (ICL})~\citep{brown2020language, min2022metaicl, chen2022meta, li2023deepinception,xiong2023dq}.
This is performed by
concatenating
a few examples (e.g., question-answer pairs)
as a prompt, and then
appending the
testing question.
However,
it is still challenging for LLMs to answer
mathematical questions
by simply prompting the question-answer pairs, as
mathematics
is more complex and often requires many steps to derive the answer.

Recently, \citet{wei2022chain}
propose \textit{chain-of-thought (CoT) prompting}, which generates explicit intermediate steps that are used to reach the answer, for LLMs.
Specifically,
each in-context example is augmented with
several thinking steps described in natural language.
A few examples are concatenated as a CoT prompt.
In inference,
the testing question
is appended to the prompt and then fed to an LLM.
The LLM is expected to imitate the in-context examples,
i.e., generating several reasoning steps before giving the answer.
CoT prompting has achieved promising performance on
mathematical reasoning tasks~\citep{wei2022chain, wang2023selfconsistency, zheng2023progressivehint, zhang2023automatic}, and many works have been proposed to improve its effectiveness~\citep{fu2023complexitybased, zheng2023progressivehint, zhou2023leasttomost, yao2023tree, pitis2023boosted} and efficiency~\citep{zhang2023automatic, kojima2022large, diao2023active, lu2022dynamic}.
	
Self-Consistency \citep{wang2023selfconsistency}
is a simple yet effective approach to improve CoT prompting.
Using temperature sampling~\citep{ackley1985learning,ficler2017controlling},
it samples a diverse set of reasoning chains
which may lead to multiple candidate answers.
The one that receives the most votes
is then chosen as the final answer.
However, our experimental results\footnote{Details are in 
Figure~\ref{fig:sc-turbo-all}
in 
Section~\ref{subsec:effect-Mf}.
}
shows that simply sampling more
reasoning paths 
may not lead to 
improvement in
testing accuracy,
particularly when the number of sampling paths is already large.
Moreover, among the failure questions of Self-Consistency,
about 60\% have at least one reasoning chain
reaching the correct answer (Table \ref{tbl:sc-fail-analysis}
in Section~\ref{apd:sc}).
Hence, the majority voting of Self-Consistency can be improved using a more reliable verifier.

\begin{figure*}[!t]
	\centering
	\includegraphics[width=0.8\textwidth]{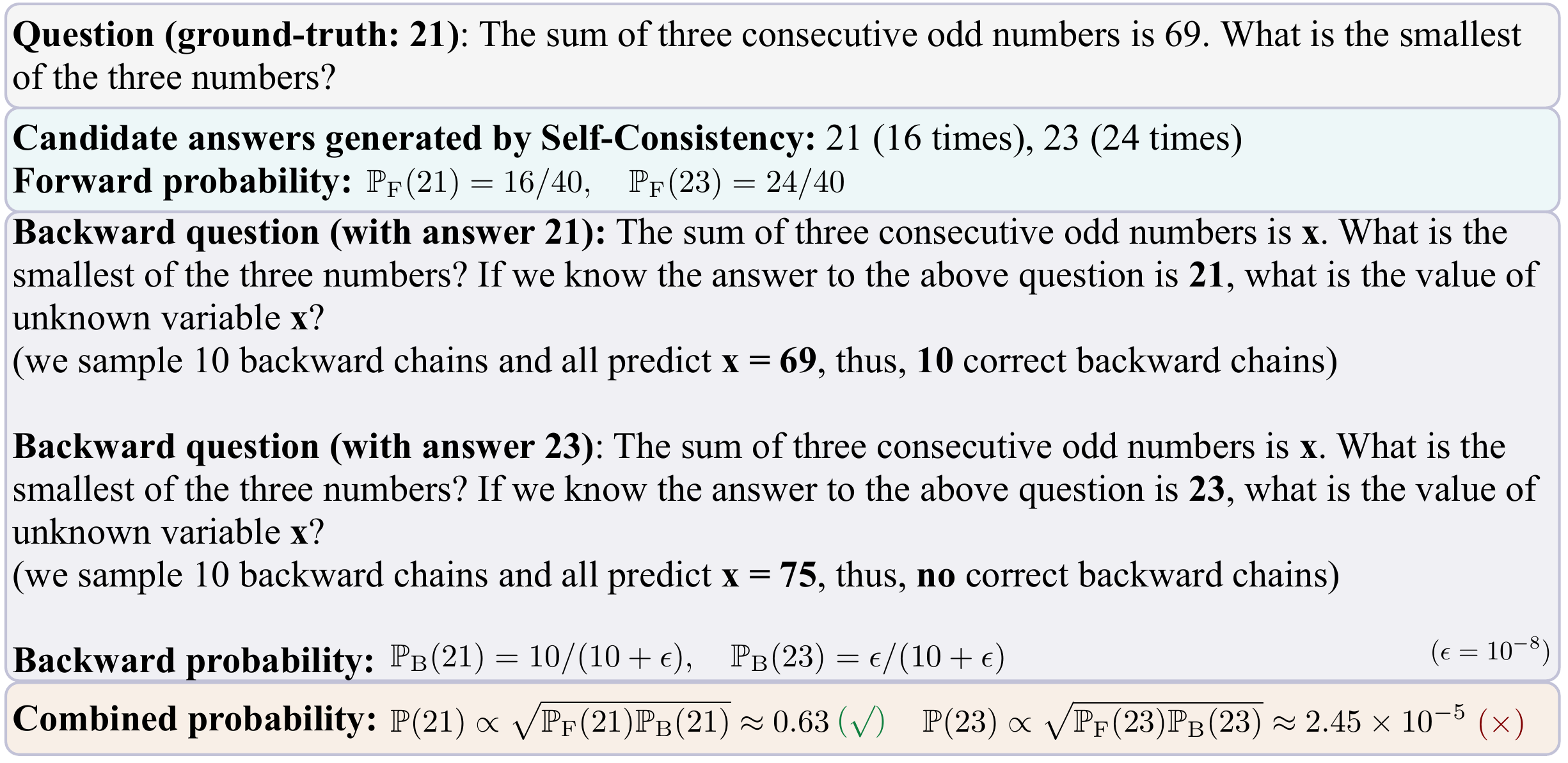}
	\vskip -.1in
	\caption{A case study for the proposed FORBA method.} 
	\label{fig:showcase}
	\vskip -.2in
\end{figure*}

We introduce backward reasoning (or backward chaining)~\citep{pettit1989backward, russell1995artificial, khot2021text, liang2021explainable, yu2023nature} to verify candidate answers. 
Figure~\ref{fig:showcase} gives a case study.
For each candidate answer  $\hat{A}_c$, 
we mask a number in the question by ``$\vx$'', and design a template ``\textit{If we know the answer to the above question is $\hat{A}_c$, what is the value of unknown variable $\vx$?}''
to form a backward question. This is then fed to the LLM to sample multiple backward reasoning chains to predict the masked number. 
As the ground-truth value of $\vx$
is known,
we can check whether the masked number is predicted correctly.
Intuitively, a correct candidate answer is more likely to help predict the masked number than wrong answers (as verified in Figure \ref{fig:backward-acc}).
Then, by defining the vote of $\hat{A}_c$ as the number of chains that predict the masked number exactly, we estimate the backward probability $\bP_{\text{B}} (\hat{A}_c )$ as the proportion of votes
$\hat{A}_c$ gets in the backward direction. When using backward reasoning alone, the prediction is $\argmax_{\hat{A}_c} \bP_\text{B}(\hat{A}_c)$.

As forward reasoning and backward reasoning are complementary, we propose a FOrward-BAckward Reasoning (FOBAR) method to combine them. By estimating the forward probability $\bP_\text{F}(\hat{A}_c)$ as the proportion of votes $\hat{A}_c$ gets in the forward direction, we propose to estimate the probability that $\hat{A}_c$ is correct (denoted by $\bP(\hat{A}_c)$) as the geometric mean of forward and backward probabilities, i.e.,
$\bP(\hat{A}_c)\propto \big(\bP_\text{F}(\hat{A}_c)\big)^{\alpha} \big(\bP_\text{B}(\hat{A}_c) \big)^{1-\alpha}$.
Extensive experiments on six data sets
and three OpenAI's LLMs (including \textit{text-davinci-003}~\citep{gpt3-5}, \textit{GPT-3.5-Turbo}~\citep{gpt3-5-turbo}, and \textit{GPT-4}~\citep{gpt4})
show that FOBAR achieves
state-of-the-art (SOTA) performance.

Our contributions are summarized as follows.
\begin{enumerate*}[(i), series = tobecont, itemjoin = \quad]
    \item We introduce backward reasoning to mathematical verification by masking a number in the original question and asking the LLM to predict the masked number when a candidate answer is provided.
    \item We propose FOBAR to combine forward and backward reasoning for verification.
    \item
    Experimental results on six standard mathematical benchmarks and three LLMs show
    that
    FOBAR
    achieves SOTA performance.
    In particular,
    FOBAR
    outperforms Self-Consistency
    which uses forward reasoning alone,
    demonstrating that combining forward and backward reasoning is better.
    Additionally,
    FOBAR outperforms Self-Verification,
    confirming  that
    using the simple template and the proposed combination is more effective.
\end{enumerate*}

\begin{figure*}[!h]
    \centering
    \includegraphics[width=.96\textwidth]{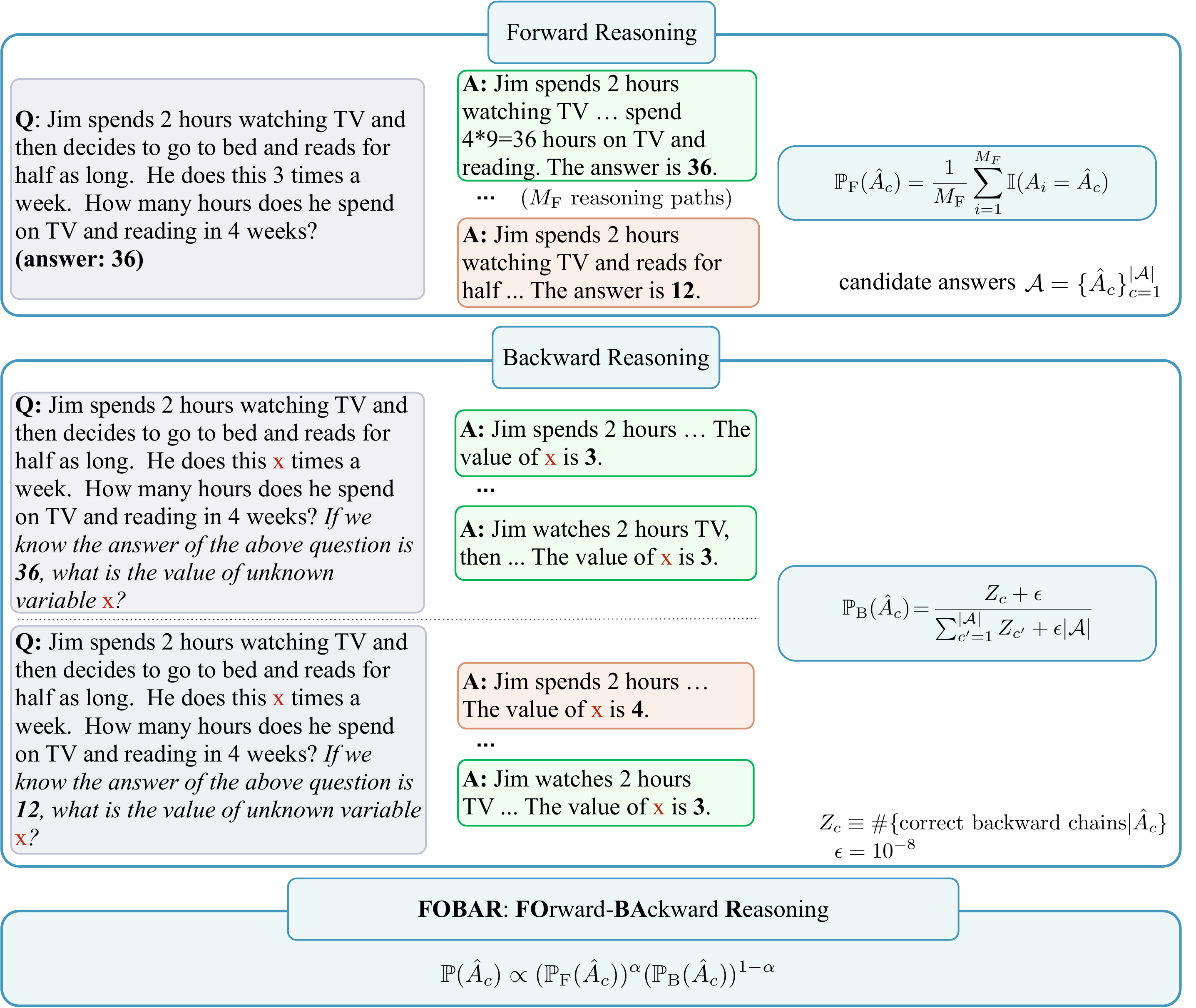}
    \vskip -.1in
    \caption{Overview of forward/backward reasoning and
        the proposed  FOBAR. The detailed procedure is shown in
        Algorithm~\ref{alg}.}
    \label{fig:overview}
    \vskip -.2in
\end{figure*}
	
\section{Related Work}
\label{sec:related-work}

\textbf{Chain-of-Thought (CoT) Prompting}.
\citeauthor{wei2022chain}
(\citeyear{wei2022chain})
propose
augmenting question-answer pairs with intermediate steps such that
the LLM can solve questions step-by-step.
Specifically,
each in-context example is a triplet $(Q^{(i)}, R^{(i)}, A^{\star(i)})$,
where $R^{(i)}$ is a reasoning chain with natural language descriptions of steps leading from the question $Q^{(i)}$ to the ground-truth answer $A^{\star(i)}$.
In inference,
a new question $Q$ is appended to the prompt:
\begin{align*}
    \!\vP_{\text{CoT}} \!&= \text{``Question: $Q^{(1)}$ \escape{n} Answer: $R^{(1)}, A^{\star(1)}$}  \nonumber \\ \!\!   \dots & \text{ Question: $Q^{(K)}$\! \escape{n} Answer: $R^{(K)}, A^{\star(K)}$''} \!\!	
\end{align*}  
and \text{``$\vP_{\text{CoT}}$ \escape{n} Question: Q \escape{n} Answer:''} is
fed to the LLM for generating both its reasoning chain $R$ and answer $A$.
CoT prompting has achieved SOTA 
performance on a wide variety of tasks
\citep{wei2022chain, kojima2022large, fu2023complexitybased, zhang2023automatic, wang2023selfconsistency, zheng2023progressivehint, zhou2023leasttomost, zhang2023multimodal,wei2024rendering}.
	
Recently, many works~\citep{fu2023complexitybased, zheng2023progressivehint, madaan2023self, paul2023refiner, shinn2023reflexion, welleck2023generating, zhou2023leasttomost, chen2023teaching, zhang2023self}
have been proposed to improve the quality of reasoning chains in CoT prompting.
ComplexCoT~\citep{fu2023complexitybased}
selects examples with more steps as in-context examples,
while
PHP~\citep{zheng2023progressivehint}
iteratively uses the previous answers as hints in prompting.
These aforementioned works can be viewed as \textit{forward reasoning}~\citep{shao2023synthetic, weng2022large}, which
starts from the question
and generates a reasoning chain
to reach the answer.
Instead of taking a single reasoning chain by greedy decoding,
Self-Consistency~\citep{wang2023selfconsistency}
samples a diverse set of chains and obtains a set of candidate answers.
The final answer is then selected
by majority voting.

\noindent\textbf{Backward Reasoning}. 
Backward reasoning (a.k.a. backward chaining)~\citep{pettit1989backward, russell1995artificial, khot2021text, liang2021explainable, yu2023nature}
starts with an answer and
works backward to
verify the sequence of steps or conditions necessary to reach this answer.
Backward reasoning is particularly useful
in domains when the answer is known,
e.g.,  in
automated theorem provers~\citep{russell1995artificial, rocktaschel2016learning, wang2020learning, kazemi2022lambada, poesia2023peano}.
Recently, Self-Verification~\citep{weng2022large} rewrites
the question with an answer into a declarative statement and then asks the LLM to predict the masked number.
RCoT~\citep{xue2023rcot}
regenerates a sentence (a sequence of tokens) in the question  conditioning  on the answer and
detects whether there is factual inconsistency in the constructed question through three complicated steps.
The complicated checking procedure may lead to inaccurate verification. 
In contrast, 
for creating backward questions,
we simply append a template to the original question without additional rewriting and reconstruction;
for verification,
the proposed FOBAR just needs to check whether the number is predicted correctly by string comparison, which is much simpler and more accurate.
Furthermore, the proposed FOBAR combines forward and backward reasoning together for verification, while Self-Verification and RCoT use backward reasoning alone.
Different from MetaMath~\cite{yu2023metamath} which uses backward reasoning to augment questions for finetuning, we focus on using backward reasoning for verification.
	

\section{Forward-Backward Reasoning for Verification}	
\label{sec:method}

\vskip -.1in
In this section, we propose the FOBAR method for verification. 
An overview is shown in 
Figure~\ref{fig:overview}.
We first consider mathematical reasoning tasks. 
A set of candidate answers 
is generated
in the forward direction, 
and we estimate each
answer's
probability
based on the votes it receives (Section~\ref{sec:method-forward}).
Next, we mask a number in the question and propose a simple template to create backward questions for verifying candidate answers (Section~\ref{sec:method-backward}).
We further propose FOBAR (Section~\ref{sec:fobar}) to combine
forward and backward reasoning.
Extension to non-mathematical tasks is discussed in Section~\ref{sec:ex-non-math}.
	
\subsection{Forward Reasoning} \label{sec:method-forward}

\textit{Forward reasoning} starts with a question and generates multiple
intermediate steps toward the answer.  Specifically, for a question $Q$, we prepend it with a base prompt $\vP_\text{F}$ (e.g., CoT prompting~\citep{wei2022chain} or ComplexCoT prompting~\citep{fu2023complexitybased}) and feed the tuple $(\vP_\text{F}, Q)$ to the LLM for generating a reasoning chain and candidate answer.  Using temperature sampling~\citep{ackley1985learning, ficler2017controlling},
we sample $M_\text{F}$
candidate reasoning chains $\{R_i\}_{i=1}^{M_\text{F}}$ and extract the corresponding candidate answers $\{A_i\}_{i=1}^{M_\text{F}}$ (see 
Figure~\ref{fig:overview},
top).
Let $\hA=\{\hat{A}_c\}_{c=1}^{|\hA|}$ be the set of answers deduplicated from
$\{A_i\}_{i=1}^{M_\text{F}}$.  Unlike greedy decoding~\citep{wei2022chain}, we may have several different candidate answers (i.e., $|\hA| > 1$).
We propose to estimate the probability that the candidate 
$\hat{A}_c\in \hA$ is correct as the proportion of votes it receives from the reasoning paths: 
\begin{equation}
    \bP_\text{F}(\hat{A}_c) = \frac{1}{M_\text{F}}\sum_{i=1}^{M_\text{F}}\bI(A_i=\hat{A}_c), \label{eq:forward}
\end{equation} 
where $\bI(\cdot)$ is the indicator function.
Choosing $\hat{A}_c$ with the largest $\bP_\text{F}(\hat{A}_c)$ corresponds to the state-of-the-art 
method of Self-Consistency~\citep{wang2023selfconsistency}. 
However,
as shown in Figure~\ref{fig:sc-turbo-all}, the performance of Self-Consistency saturates when $M_\text{F}$ is sufficiently large.
Thus, simply sampling more reasoning paths brings negligible performance improvement.

\subsection{Backward Reasoning}	\label{sec:method-backward}

In backward reasoning,
we mask a number contained in the question and
ask
the LLM
to predict the masked number by using
a provided candidate answer.
Specifically, suppose
that question $Q$ involves $N_Q$ numbers
$\{\text{num}^{(n)}\}_{n=1}^{N_Q}$.
We replace each of them one by one with
$\vx$.
The resultant masked question
$\hat{Q}^{(n)}$
is then concatenated with
the following template,
which contains a candidate answer $\hat{A}_c \in \hA$.
\begin{prompt*}{}{}
    \centering
    {\small \!\!\!\!$\hT(\hat{A}_c)$ = \textit{If we know the answer to the above question is \\ \!\!\!\!\!\!\!\!\!\!\!\!\!\!\!\!\!\!\!\!\!\!\!\!
    		$\{\hat{A}_c\}$, what is the value of unknown variable {\normalfont x}?}}
\end{prompt*}

Each $(\hat{Q}^{(n)},\hT(\hat{A}_c))$ pair is called a
\textit{backward question}.
In total, we obtain $N_Q$
backward questions. 
Some examples of backward questions are shown in Example~\ref{exmp:inv-q} of Appendix~\ref{sec:exmp-back}.
Note that Self-Verification~\citep{weng2022large} needs the assistance of an LLM to
rewrite a (question, answer) pair into a declarative statement.\footnote{For example, ``How many hours does he spend on TV and reading in 4 weeks?'' with a candidate answer of 36 is rewritten to ``He spends 36 hours on TV and reading in 4 weeks''.} 
In contrast, the proposed template 
is simpler and avoids possible
mistakes
(an example illustrating
Self-Verification's
rewriting mistakes  
is shown in Appendix~\ref{appendix:fail-sv}).

To predict the masked number,
we prepend the backward
question with a prompt $\vP_{\text{B}}$, which consists of several (backward)
question-answer demos with reasoning chains.  An example
question-answer demo is shown in Example \ref{exmp:backward} of Appendix \ref{sec:exmp-back}. 
We feed each of $(\vP_\text{B}, \hat{Q}^{(n)}, \hT(\hat{A}_c))$
(where $n=1,\dots,N_Q$) 
to the LLM, which then
imitates the in-context examples in
$\vP_\text{B}$
and
generates a reasoning chain for the prediction of the masked number.
We sample
$M_{B}$  such reasoning chains with predictions $\{\widehat{\text{num}}_{c,b}^{(n)}\}_{b=1}^{M_{\text{B}}}$ (see
Figure~\ref{fig:overview},
middle).
For each candidate answer $\hat{A}_c$, we count the number of times that
the masked number
is exactly predicted: 
\begin{equation}
    Z_c = \sum_{n=1}^{N_Q} \sum_{b=1}^{M_{\text{B}}}
    \bI(\widehat{\text{num}}_{c,b}^{(n)}= \text{num}^{(n)}). \label{eq:correct-backward}
\end{equation}
The probability that  candidate answer
$\hat{A}_c$ is correct
is estimated as  
\begin{equation}
\bP_\text{B}(\hat{A}_c) = \frac{Z_c + \epsilon}{\sum_{c'=1}^{|\hA|}Z_{c'} + \epsilon |\hA|}, \label{eq:backward}
\end{equation}  
where $\epsilon=10^{-8}$ is a
small positive constant to avoid division by zero.
One can simply choose $\hat{A}_c$
with the largest
$\bP_\text{B}(\hat{A}_c)$
as the prediction.
A more effective method,
as will be shown in Section~\ref{sec:fobar},
is to combine the probabilities obtained from
forward and backward reasoning.
	
\subsection{FOBAR
    {\small (\textbf{FO}rward and \textbf{BA}ckward \textbf{R}easoning)}}
\label{sec:fobar}

As forward and backward reasoning
are complementary (i.e., backward reasoning may succeed in the cases where forward reasoning fails, and vice versa, as shown in Examples \ref{exmp:fw-w-bw-r} and
\ref{exmp:fw-r-bw-w} in Appendix~\ref{apd:example-bw-fw}), we propose to combine them for verification.  Intuitively, a
candidate answer is likely to be correct when it receives many votes in
forward reasoning and also helps the LLM to predict the masked numbers in backward reasoning.
We estimate the probability that $\hat{A}_c$ is correct as  
\begin{align}
\!\bP(\hat{A}_c)\! \propto\! \big(\bP_\text{F}(\hat{A}_c)\big)^{\alpha} \big(\bP_\text{B}(\hat{A}_c) \big)^{1-\alpha}, \label{eq:fobar}
\end{align} 
with weight 
$\alpha\in [0,1]$ (see
Figure~\ref{fig:overview},
bottom).
When $\alpha =1$,
it reduces to Self-Consistency~\citep{wang2023selfconsistency}; When $\alpha$ equals $0$,
it reduces to backward reasoning for verification.
In the experiments, we combine the forward and backward probabilities by
the geometric mean
(i.e., $\alpha= 0.5$) since we expect the final candidate answer to have non-negligible probabilities in both forward and backward directions.
Finally, we select the answer as
$\argmax_{\hat{A}_c \in \hA} \bP(\hat{A}_c)$.
The whole procedure is shown in Algorithm \ref{alg}.
As all the probability calculations are simple, the additional computation cost of Algorithm \ref{alg} is negligible.

Compared with training an LLM as verifier~\citep{cobbe2021training},
which is computationally expensive and labor-intensive in collecting extra annotation data,
FOBAR is training-free (thus, no additional data collection) and more effective in verification (Table~\ref{table:baseline} in Appendix~\ref{sec:train-verfier}).
The proposed backward reasoning can be combined with other forward reasoning methods such as 
step-by-step verification proposed by \citet{ling2023deductive} (Table~\ref{table:ded-verx} in Appendix~\ref{apd:step-step-forward}).

\begin{algorithm}[!h]
    \caption{FOBAR.}
    \label{alg}
    \begin{algorithmic}[1]
        \Require  number of
        reasoning chains
        $M_\text{F}$ and $M_\text{B}$, prompts $\vP_\text{F}$ and $\vP_\text{B}$; $\epsilon=10^{-8}$; $\alpha=0.5$; 
        \State \textbf{Input}: a  question $Q$ with $N_Q$ numbers;
        \State feed $(\vP_\text{F}, Q)$ to LLM, sample $M_\text{F}$ reasoning chains with candidate answers $\{A_i\}_{i=1}^{M_\text{F}}$;
        \State deduplicate $\{A_i\}_{i=1}^{M_\text{F}}$ to $\hA=\{\hat{A}_c\}_{c=1}^{|\hA|}$;
        \State compute $\bP_\text{F}(\hat{A}_c)$ by Eq.~\eqref{eq:forward} for $\hat{A}_c \!\in\! \hA$;
        \For{$\hat{A}_c\in \hA$}
        \For{$n=1,\dots,N_Q$}
        \State create $\hat{Q}^{(n)}$ by masking the $n$th  \Statex \quad \quad \quad  number $\text{num}^{(n)}$ in $Q$;
        \State feed $(\vP_\text{B}, \hat{Q}^{(n)}, \hT(\hat{A}_c))$ to LLM;
        \State sample $M_\text{B}$ predictions $\{\widehat{\text{num}}_{c, b}^{(n)}\}_{b=1}^{M_\text{B}}$;
        \EndFor
        \State compute $Z_c$ by Eq.~\eqref{eq:correct-backward};\!\!\!\!\!\! \label{alg-step:correctness}
        \EndFor
        \State compute $\bP_\text{B}(\hat{A}_c)$ by Eq.~\eqref{eq:backward} for $\hat{A}_c \!\in\! \hA$;
        \State compute $\bP(\hat{A}_c)$ by Eq.~\eqref{eq:fobar}  for $\hat{A}_c \in \hA$; \\
        \Return $\argmax_{\hat{A}_c \in \hA} \bP(\hat{A}_c)$.
    \end{algorithmic}
\end{algorithm}

\subsection{Extension to Non-Mathematical Tasks}
\label{sec:ex-non-math}

In mathematical questions,
numbers are the most informative words.
For non-mathematical tasks, we can analogously
mask an informative word and 
ask the LLM to guess the masked word given a candidate answer.
For example, consider
the following question-answer pair 
from the \textit{Last Letter Concatenation} task~\citep{wei2022chain, zhou2023leasttomost}:
``Take the last letters of each word in `\textit{Whitney Erika Tj Benito}' and concatenate them'' with ground-truth answer ``\textit{yajo}''. We can mask one of the four words (e.g., ``\textit{Erika}'').
Given a candidate answer $\hat{A}_c$, 
we create a backward question as
``Take the last letters of each word in `\textit{Whitney \_\_\_ Tj Benito}' and concatenate them. If we know the answer to the above question is $\hat{A}_c$, which is the word at the blank, \textit{Erika} or \textit{Dqhjz}'',
where ``\textit{Dqhjz}'' is obtained by shifting each letter of ``\textit{Erika}''.
The LLM is more likely to choose ``\textit{Erika}''  if the second letter in $\hat{A}_c$ is ``\textit{a}''.


\section{Experiments}
\label{sec:expt}

\subsection{Setup}
\label{sec:expt-setup}

\paragraph{Datasets.}
Experiments are conducted on six benchmark 
mathematical data sets which are commonly used in evaluating CoT reasoning ability~\citep{zheng2023progressivehint, wang2023selfconsistency}:
\begin{enumerate*}[(i), series = tobecont, itemjoin = \quad]
    \item
    \textit{AddSub}~\citep{hosseini2014learning},
    \item
    \textit{MultiArith}~\citep{roy2015solving},
    \item
    \textit{SingleEQ}~\citep{koncel2015parsing},
    \item
    \textit{SVAMP}~\citep{patel2021nlp},
    \item
    \textit{GSM8K}~\citep{cobbe2021training},
    \item
    \textit{AQuA}~\citep{ling2017program}.
\end{enumerate*}
Some statistics and example question-answer pairs are shown in Table \ref{table:data sets}
in Appendix~\ref{appendix:dataset}.
Questions in \textit{AddSub} and \textit{SingleEQ} are easier and do not need multi-step calculations. Questions in the other data sets
are more challenging as many steps are required.

\paragraph{Baselines.}
We compare the proposed FOBAR with
\begin{enumerate*}[(i), series = tobecont, itemjoin = \quad]
    \item
    In-Context Learning (ICL) using question-answer pairs as demonstrations~\citep{brown2020language},
    and recent CoT prompting methods, including:
    \item
    CoT prompting~\citep{wei2022chain};
    \item
    ComplexCoT prompting~\citep{fu2023complexitybased} which
    selects demonstrations with complex reasoning steps;
    \item RE2~\citep{xu2023re} which re-reads the question in the prompt;
    \item
    PHP~\citep{zheng2023progressivehint} which
    iteratively uses the previous answers as hints in designing prompts;
    \item RCoT~\citep{xue2023rcot} which reconstructs the question based on the candidate answer and checks the factual inconsistency for verification;
    \item RCI~\citep{kim2023language} which recursively criticizes and improves its previous output;
    \item
    Self-Consistency~\citep{wang2023selfconsistency}, which samples multiple
    reasoning chains and selects the answer by majority voting;
    \item Self-Verification~\citep{weng2022large}, which
    chooses the top-2 candidate answers obtained from Self-Consistency and
    re-ranks them based on the verification scores computed in the backward procedure.
\end{enumerate*}

Following~\citet{zheng2023progressivehint},
we experiment with three LLMs:
\begin{enumerate*}[(i), series = tobecont, itemjoin = \quad]
    \item
    \textit{text-davinci-003}~\citep{gpt3-5},
    \item
    \textit{GPT-3.5-Turbo}~\citep{gpt3-5-turbo},
    and
    \item
    \textit{GPT-4}~\citep{gpt4}.
\end{enumerate*}
\textit{GPT-3.5-Turbo} and \textit{GPT-4}
are more powerful than \textit{text-davinci-003}.
The proposed FOBAR is general and
can be integrated into any prompting method.
Here,
we choose
the CoT prompting
and
ComplexCoT prompting
as base prompts as in~\citet{zheng2023progressivehint}.

\noindent
\textbf{Implementation Details.}
Following~\citep{wang2023selfconsistency, zhou2023leasttomost, zheng2023progressivehint},
the temperature for sampling is 0.7
for both forward and backward reasoning.
The $\alpha$ in Eq.~(\ref{eq:fobar}) is set to 0.5.
For \textit{text-davinci-003},
$M_\text{F}$ is 40 as in~\citep{wang2023selfconsistency, zheng2023progressivehint};
whereas the more powerful LLMs
(\textit{GPT-3.5-Turbo} and \textit{GPT-4})
use a smaller
$M_\text{F}$ (i.e., 10).
$M_\text{B}$ is set to 8 for all three LLMs.
We do not repeat the experiments using different seeds as querying OpenAI's LLMs is costly, 
which is a standard protocol in CoT-based research~\citep{fu2023complexitybased, wang2023selfconsistency, zhou2023leasttomost}.
The number of forward chains is identical for Self-Consistency, Self-Verification, and FOBAR,
while the number of backward chains is identical for Self-Verification and FOBAR

\subsection{Main Results}
\label{sec:expt-results}

Table~\ref{table:main}
shows the testing accuracies.
As can be seen,
for all three LLMs,
FOBAR with ComplexCoT prompting
achieves the highest average accuracy,
showing that FOBAR is effective in verifying candidate answers. 
This new finding suggests that verification is a promising direction to improve the performance of CoT-based methods.
When using CoT as the base prompt,
FOBAR outperforms Self-Consistency most of the time,
demonstrating that
combining forward and backward reasoning is better than using forward reasoning alone.
Furthermore, FOBAR performs better than Self-Verification on almost all datasets,
demonstrating that using the proposed simple template in backward reasoning and the proposed combination is more effective in verification.
FOBAR
(with either CoT or ComplexCoT)
on \textit{GPT-4} achieves the highest 
average
accuracy, as \textit{GPT-4} is currently the SOTA LLM.
Moreover,
for all three LLMs,
FOBAR using ComplexCoT as base prompt achieves higher accuracy
than
using CoT on average, which is consistent with observations in~\citep{fu2023complexitybased, zheng2023progressivehint} that ComplexCoT is better than CoT.

\begin{table*}[!t]
\centering
\vskip -.2in
\caption{Testing accuracies (\%) on six data sets using three LLMs. For each LLM,
    methods are grouped according to the base prompt they used. The best in each group is in \textbf{bold}.
    Results with $^\dagger$ are from the original publications.
    ``--'' means that the result is not reported in the original publication.
}
\vskip -.15in
\label{table:main}
\resizebox{.95\textwidth}{!}{
    \renewcommand{\arraystretch}{1.05}
    \begin{NiceTabular}{c|c|l@{\hspace{0cm}}ccccccc}
        \CodeBefore
        \rectanglecolor{orange3!50}{3-2}{8-2}
        \rectanglecolor{blue3!50}{9-2}{13-2}
        \rectanglecolor{orange3!50}{15-2}{19-2}
        \rectanglecolor{blue3!50}{20-2}{26-2}
        \rectanglecolor{orange3!50}{28-2}{31-2}
        \rectanglecolor{blue3!50}{32-2}{36-2}
        \rectanglecolor{Gray}{8-3}{8-10}
        \rectanglecolor{Gray}{13-3}{13-10}
        \rectanglecolor{Gray}{19-3}{19-10}
        \rectanglecolor{Gray}{26-3}{26-10}
        \rectanglecolor{Gray}{31-3}{31-10}
        \rectanglecolor{Gray}{36-3}{36-10}
        \Body
        \arrayrulecolor{black}\specialrule{1.3pt}{.3\jot}{0.3pc}
        & & & \textit{AddSub} & \textit{MultiArith} & \textit{SingleEQ} & \textit{SVAMP} & \textit{GSM8K} & \textit{AQuA} & Average \\
        \arrayrulecolor{black}\specialrule{1.3pt}{.3\jot}{0.3pc}
        \multirow{13}{*}{\STAB{\rotatebox[origin=c]{90}{\textit{text-davinci-003}}}}
        & &
        ICL~\citep{brown2020language} &  90.4 &  37.6 &84.3 &69.1&16.9& 29.1 & 54.5\\
        \cmidrule{2-10}
        & \multirow{6}{*}{\STAB{\rotatebox[origin=c]{90}{CoT}}} 
        &CoT~\citep{wei2022chain} & 91.4 & 93.6 & 92.7 & 79.5 & 55.8 & 46.5 & 76.6 \\
        && PHP$^\dagger$~\citep{zheng2023progressivehint} & 91.1 & 94.0 & 93.5 & 81.3 & 57.5 & 44.4 & 77.0 \\
        &&RE2$^\dagger$~\citep{xu2023re} & 91.7 & 93.3 & 93.3 & 81.0 & 61.6 & 44.5 & 77.6 \\
        &&Self-Consistency~\citep{wang2023selfconsistency} & 91.7 &   95.9 & 94.5 & 83.1 &67.9& \textbf{55.1 }& 81.4 \\
        &&Self-Verification~\citep{weng2022large} & 87.4 & 95.3 & 92.9 & 82.2 & 59.8 & 37.4 & 75.8 \\
        &&FOBAR   &{ \textbf{91.9}} & {\textbf{100.0}} & {\textbf{96.1} }& \textbf{86.8} & \textbf{70.8} & \textbf{55.1} & \textbf{83.5}  \\
        \cmidrule{2-10}
        & \multirow{5}{*}{\STAB{\rotatebox[origin=c]{90}{ComplexCoT}}} 
        & ComplexCoT~\citep{fu2023complexitybased}  & 88.9 & 95.3 & 93.7 & 78.0 & 67.7 & 48.8&78.7\\
        &&PHP$^\dagger$~\citep{zheng2023progressivehint} & \textbf{91.6} & 96.6 & 95.0 & 83.7 & 68.4 & 53.1 & 81.4 \\
        &&Self-Consistency~\citep{wang2023selfconsistency} & 89.4 &98.5  & 91.1& 82.7 & {\textbf{79.1}} & {\textbf{58.7}} & 83.2 \\
        &&Self-Verification~\citep{weng2022large} & 89.9 & 95.5 & 94.1 & 80.1 & 72.0 & 38.2 &78.3 \\
        &&FOBAR   & 90.6 &{ \textbf{100.0} }& \textbf{95.3} &{ \textbf{87.0}} & 78.7 & {\textbf{58.7}} & {\textbf{85.0}} \\
        \arrayrulecolor{black}\specialrule{1.3pt}{.3\jot}{0.3pc}
        \multirow{14}{*}{\STAB{\rotatebox[origin=c]{90}{\textit{GPT-3.5-Turbo}}}} & &
        ICL~\citep{brown2020language} & 88.6 & 87.6 & 88.8 & 80.6 & 32.2 & 31.1 & 68.2\\
        \cmidrule{2-10} 
        & \multirow{5}{*}{\STAB{\rotatebox[origin=c]{90}{CoT}}} 
        &CoT~\citep{wei2022chain} & 89.4 & 97.9   & 92.9 & 84.2 & 77.2 & 54.3 & 82.7 \\
        &&RE2$^\dagger$~\citep{xu2023re} & 89.9 & 96.5 & {\textbf{95.3}} & 80.0 & 80.6 & 58.3 & 83.4 \\
        &&Self-Consistency~\citep{wang2023selfconsistency}&{ \textbf{90.6}} & 98.6 & 93.1 & 86.4 & 81.9 & {\textbf{62.6}} & 85.5\\
        &&Self-Verification~\citep{weng2022large} & 90.4 & 97.4 & 92.9 & 83.1 & 74.9 & 60.6 & 83.2 \\
        && FOBAR & 89.4 & \textbf{99.3} & 94.5 & {\textbf{88.9}} & \textbf{85.1} & {\textbf{62.6}} & {\textbf{86.6}} \\
        \cmidrule{2-10}
        & \multirow{6}{*}{\STAB{\rotatebox[origin=c]{90}{ComplexCoT}}} 
        &Complex CoT~\citep{fu2023complexitybased} & 87.9 & 98.3 & \textbf{94.5} & 81.1 & 80.7 &  59.1 & 83.6 \\
        &&RCoT$^\dagger$~\citep{xue2023rcot} & 88.2 & -- & 93.0 & 84.9 & 84.6 & 53.3 & -- \\
        &&PHP$^\dagger$~\citep{zheng2023progressivehint} & 85.3& 98.0 & 92.9 & 83.1 & 85.1 & 60.6  & 84.2 \\
        && RCI$^\dagger$~\citep{kim2023language} & \textbf{90.6} & 99.21 & 93.7 & 87.4 & 84.3 & -- & -- \\
        &&Self-Consistency~\citep{wang2023selfconsistency} & 88.1 &  98.8 & \textbf{94.5 }& 85.0&86.4 & 63.0 &  86.0 \\
        &&Self-Verification~\citep{weng2022large}  & 87.9 & 96.6 & 93.3 & 81.0 & 78.2 & 61.4 & 83.1 \\
        &&FOBAR & {88.4} & {\textbf{99.8} }& 94.3 & \textbf{88.5} & {\textbf{87.4}} & {\textbf{63.4}} & {\textbf{87.0}} \\
        \arrayrulecolor{black}\specialrule{1.3pt}{.3\jot}{0.3pc}
        \multirow{10}{*}{\STAB{\rotatebox[origin=c]{90}{\textit{GPT-4}}}}
        && ICL ~\citep{brown2020language} & 92.1 & 98.6 & 94.3 & 90.9 & 48.5 & 48.0 & 78.7 \\
        \cmidrule{2-10}
        & \multirow{4}{*}{\STAB{\rotatebox[origin=c]{90}{CoT}}} 
        & CoT~\citep{wei2022chain} &  {\textbf{92.7}} & {\textbf{ 99.0}} & 95.7 & 92.9 &93.4&69.7 & 90.6 \\
        && Self-Consistency~\citep{wang2023selfconsistency}& 92.2 &  {\textbf{99.0}} & 95.9 & 93.3 & 94.8 & \textbf{71.3} & 91.1\\
        && Self-Verification~\citep{weng2022large} & {\textbf{92.7}} & {\textbf{99.0}} & 95.7 & 93.1 & 93.7 & 70.1 & 90.7 \\
        && FOBAR & 92.4 & {\textbf{99.0}} & {\textbf{96.1}}& \textbf{94.1} & \textbf{95.4} & \textbf{71.3} & \textbf{91.4} \\
        \cmidrule{2-10}
        & \multirow{5}{*}{\STAB{\rotatebox[origin=c]{90}{ComplexCoT}}} 
        & Complex CoT~\citep{fu2023complexitybased} & \textbf{91.9} & 98.3 & 94.5 & 92.4 & 95.1 & 72.4 & 90.8\\
        && PHP$^\dagger$~\citep{zheng2023progressivehint} &  89.6 & 98.1 & 93.1 & 91.9 & 95.5 & {\textbf{79.9}}  & 91.3\\
        && Self-Consistency~\citep{wang2023selfconsistency} &  91.4 & 98.5 & \textbf{94.7} & 93.4  & 96.2 & 75.2 & 91.6 \\
        && Self-Verification~\citep{weng2022large} & 91.6 & 98.5 & \textbf{94.7} & 93.0 & 95.7 & 75.6 & 91.5\\
        && FOBAR & \textbf{91.9} & \textbf{98.6} & \textbf{94.7} & {\textbf{94.4}} & {\textbf{96.4}} & 75.2 & {\textbf{91.9}} \\
        \arrayrulecolor{black}\specialrule{1.3pt}{.3\jot}{0.3pc}
    \end{NiceTabular}
}
\vskip -.2in
\end{table*}

\subsection{Combining
Forward and Backward Probabilities}
\label{sec:abl-alpha}

\begin{figure}[!h]
    \centering
     \vskip -.1in	
    \!\!\!
    \subfigure[\label{fig:all-complexcot-003}\textit{text-davinci-003}.]{\includegraphics[width=0.16\textwidth]{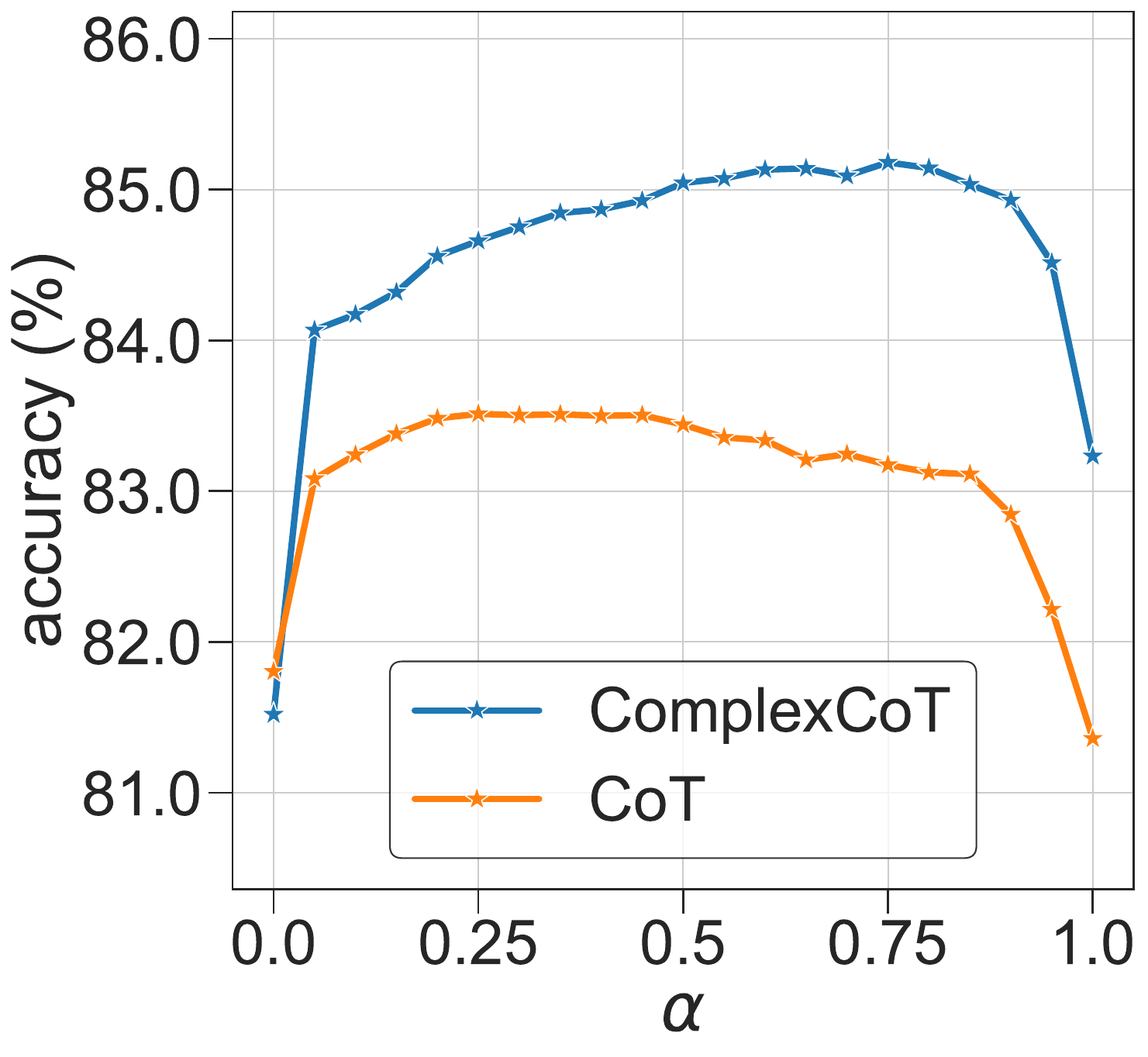}} 
    \subfigure[\label{fig:all-complexcot-3.5turbo}\!\! \textit{GPT-3.5-Turbo}.]{\includegraphics[width=0.16\textwidth]{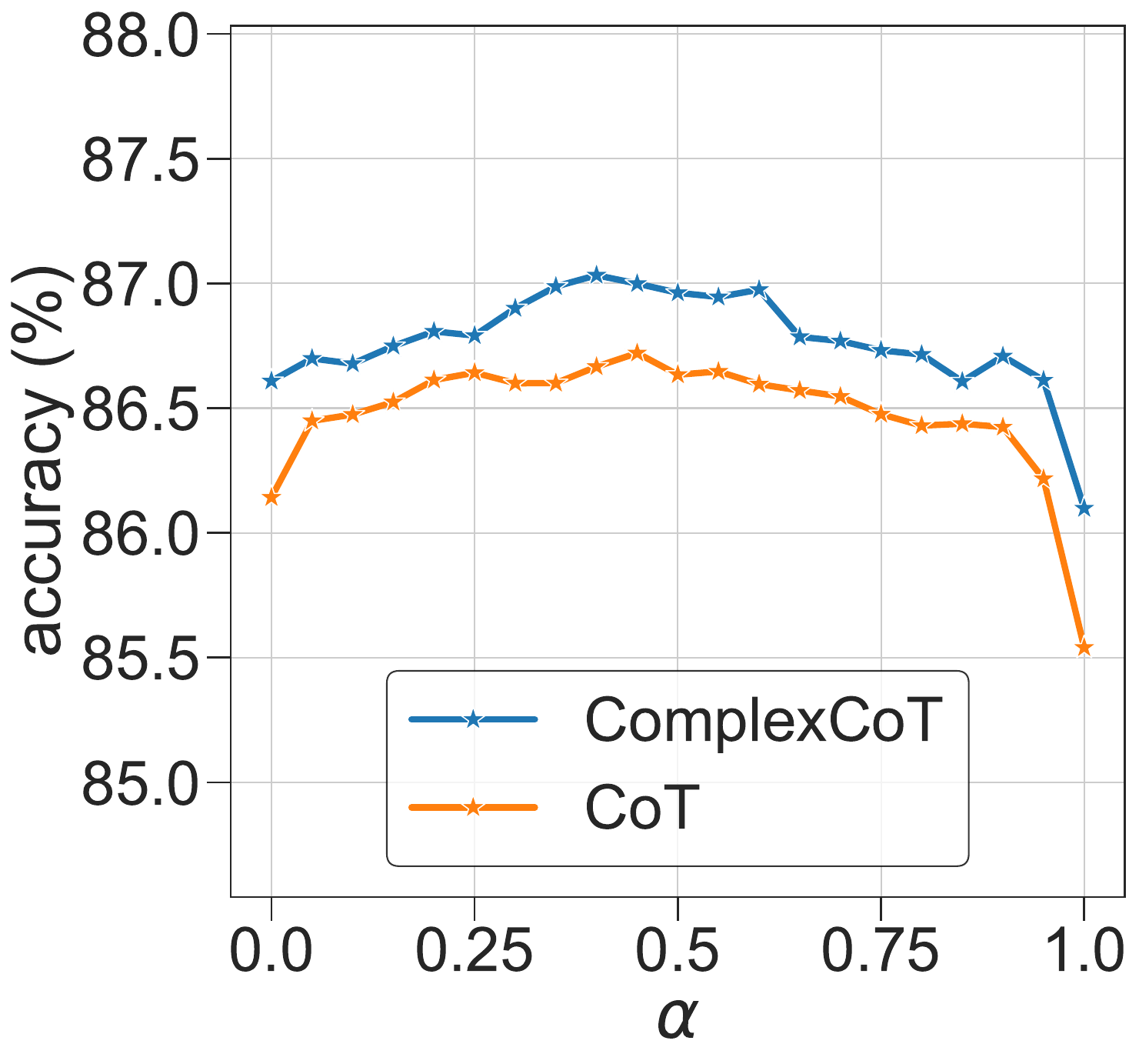}} \!\!
    \subfigure[\label{fig:all-complexcot-gpt4}\!\! \textit{GPT-4}.]{\includegraphics[width=0.16\textwidth]{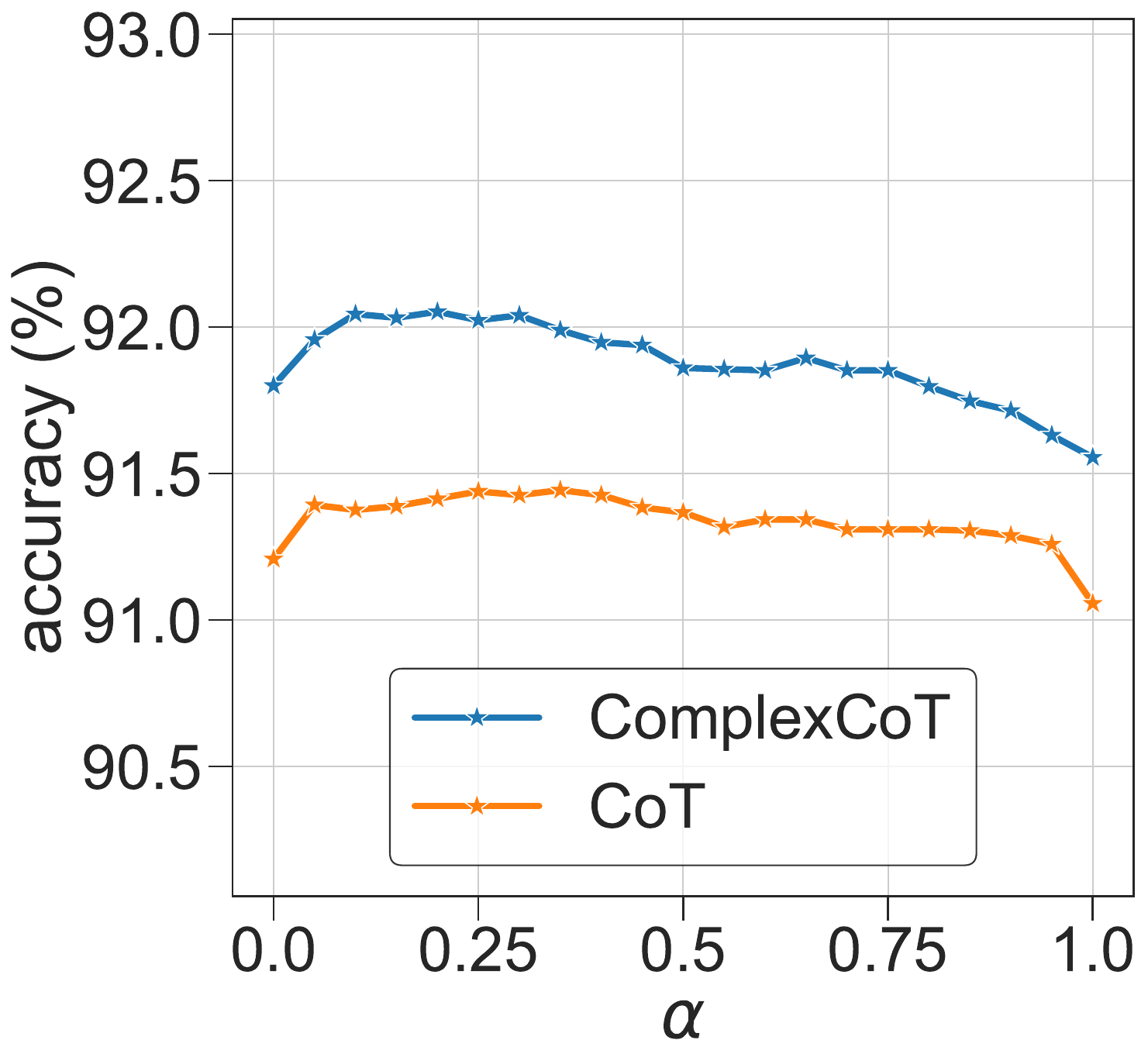}} \!\!\!
    \vskip -.18in
    \caption{Testing accuracy 
        (averaged over
        the six data sets)
        of FOBAR w.r.t. $\alpha$.
        \label{fig:albation-alpha-turbo-all}}
\end{figure}

In this experiment, we study how the combination weight $\alpha$ in Eq.~\eqref{eq:fobar}
affects performance.
Figure~\ref{fig:albation-alpha-turbo-all}
shows the
testing accuracies
(averaged over the six data sets)
with $\alpha\in [0,1]$ using the three LLMs.
As can be seen,
FOBAR is insensitive
to $\alpha$ over a large range
for all three LLMs.
In the sequel,
we use $\alpha=0.5$, which corresponds to the geometric
mean of the forward and backward probabilities.

\begin{figure}[!h]
    \centering
    \!\!\!\!
    \subfigure[\label{fig:abl-backward-003}\textit{text-davinci-003}.]{\includegraphics[width=0.16\textwidth]{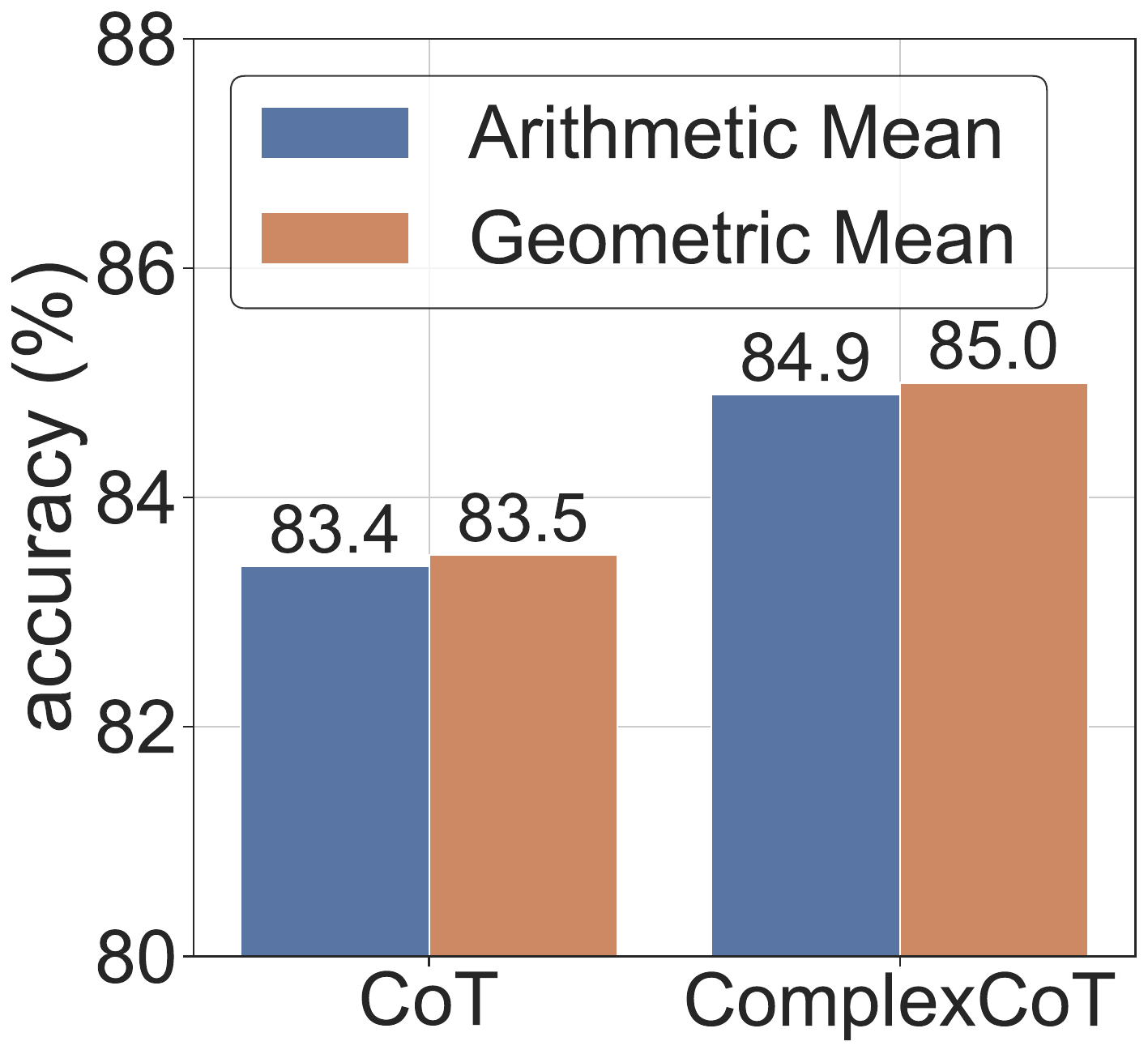}}
    \subfigure[\label{fig:abl-backward-turbo}\textit{GPT-3.5-Turbo}.]{\includegraphics[width=0.16\textwidth]{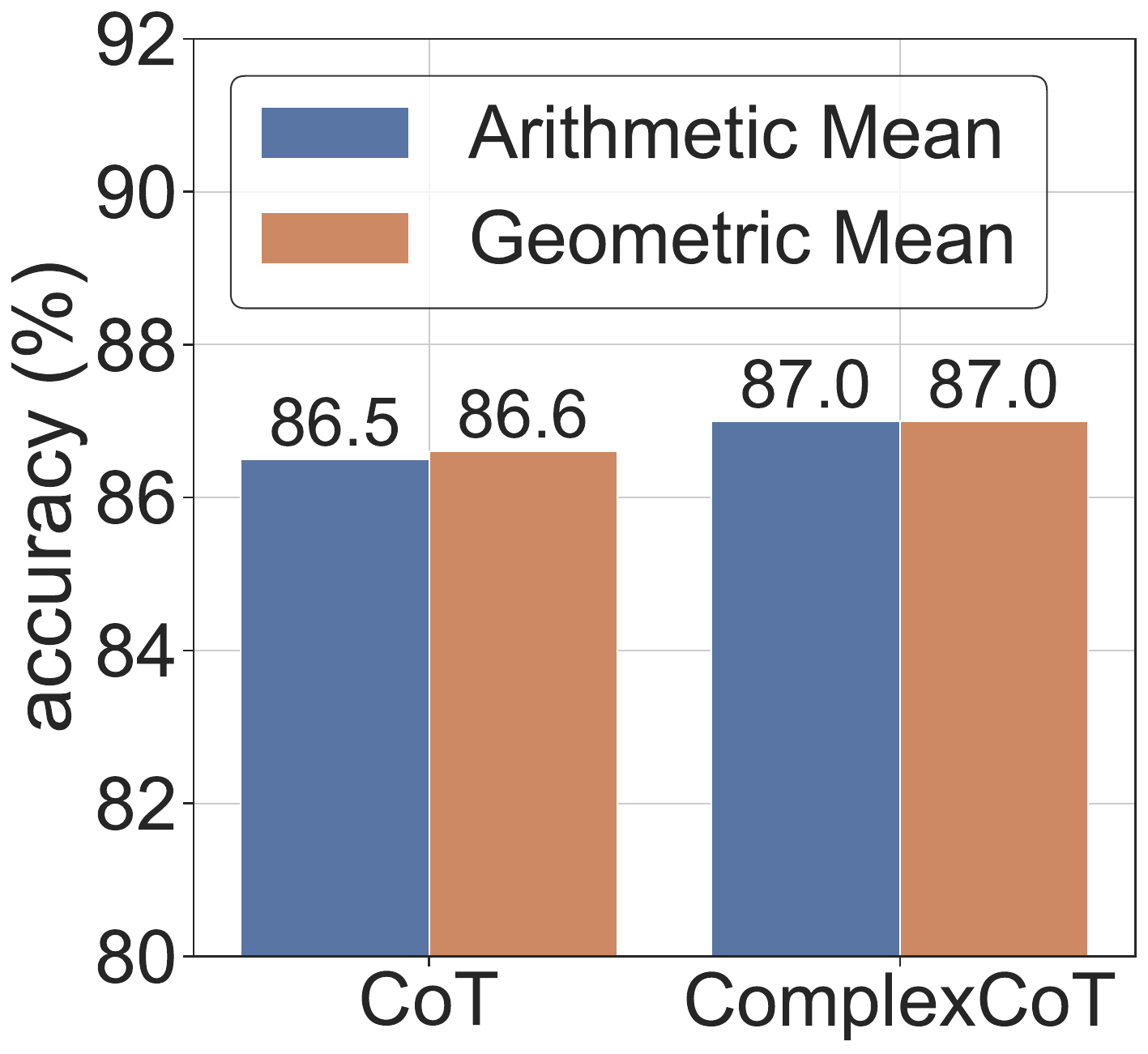}}
    \!
    \subfigure[\label{fig:abl-backward-gpt4}\textit{GPT-4}.]{\includegraphics[width=0.16\textwidth]{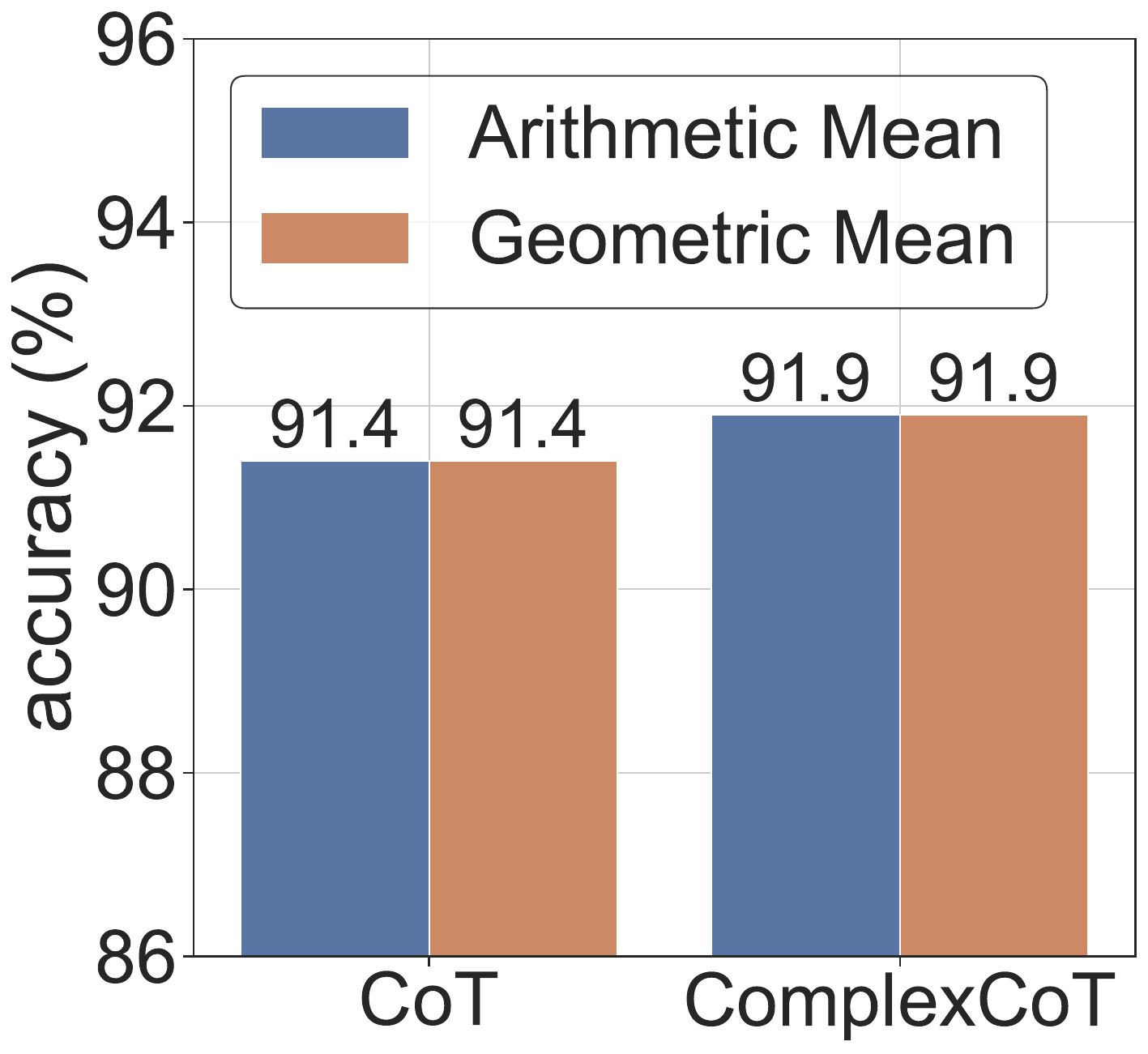}}
    \!\!\!\!
    \vskip -.18in
    \caption{Testing accuracy of FOBAR
        (averaged
        over the six data sets)
        with geometric/arithmetic mean of
        forward and backward probabilities.}
    \label{fig:albation-comb-all}
    \vskip -.25in
\end{figure}

Alternatively, one can 
combine the forward and backward probabilities
by the arithmetic mean, i.e.,
$\bP(\hat{A}_c) = \frac{1}{2}\big(\bP_\text{F}(\hat{A}_c) + \bP_\text{B}(\hat{A}_c)\big)$.
Figure~\ref{fig:albation-comb-all}
shows the testing accuracies for the three LLMs.
As shown,
the arithmetic mean has comparable performance as the geometric mean.
Hence,
Figures~\ref{fig:albation-alpha-turbo-all} and
\ref{fig:albation-comb-all} together
suggest that FOBAR is robust to the combination of
forward and backward probabilities.

\subsection{Usefulness of Forward and Backward Reasoning}

We perform an ablation study on forward (FO) and backward (BA) reasoning.
We consider the four combinations:
\begin{enumerate*}[(i), series = tobecont, itemjoin = \quad]
	\item using
	neither forward nor backward reasoning (which reduces to greedy
	decoding~\citep{wei2022chain});
	\item
	use only forward reasoning (i.e., Self-Consistency);
	\item use only
	backward reasoning in selecting answers (i.e., $\alpha=0$ in Algorithm \ref{alg});
	\item use
	both forward and backward reasoning (i.e., the proposed FOBAR).
\end{enumerate*}
Table~\ref{table:effect-of-comb}
shows the
testing accuracies (averaged
over the six data sets) for the three LLMs.
As can be seen,
in all the settings,
using forward or backward reasoning is consistently better
than using neither of them.
Moreover,
combining forward and backward
reasoning is always the best.
Examples \ref{exmp:fw-w-bw-r} and
\ref{exmp:fw-r-bw-w} in Appendix~\ref{apd:example-bw-fw}
show that FOBAR is able to rectify some failure cases of forward and backward reasoning, respectively.

\begin{table}[!t]
	\centering
	\caption{Average testing accuracies (\%) with different combinations of forward (FO) and backward (BA) reasoning.}
	\vskip -.15in
	\renewcommand{\arraystretch}{1.1}
	\label{table:effect-of-comb}
	\resizebox{.48\textwidth}{!}{
		\begin{NiceTabular}{ccc|ccc}
			\CodeBefore
                \rectanglecolor{orange3!50}{2-1}{5-1}
			\rectanglecolor{blue3!50}{6-1}{9-1}
			\rectanglecolor{Gray}{5-2}{5-6}
			\rectanglecolor{Gray}{9-2}{9-6}
			\Body
			\toprule
			& FO & BA & \textit{text-davinci-003} & \textit{GPT-3.5-Turbo } & \textit{GPT-4}\\
			\midrule
			\multirow{4}{*}{\STAB{\rotatebox[origin=c]{90}{CoT}}}  & 
			\xmark & \xmark & 76.6 & 82.7 & 90.6  \\
			&	\cmark & \xmark & 81.4 & 85.5 & 91.1 \\
			&	\xmark & \cmark  & 82.1 &  86.2 & 91.2 \\ 
			&	\cmark & \cmark  & \textbf{83.5}& \textbf{86.6} & \textbf{91.4}\\
			\midrule
			\multirow{4}{*}{\STAB{\rotatebox[origin=c]{90}{ComplexCoT}}}  & 
			\xmark & \xmark & 78.7 & 83.6 & 90.8\\
			&	\cmark & \xmark & 83.2 & 86.0 &  91.6\\
			&	\xmark & \cmark  & 81.3 & 86.3 &  91.8 \\
			&	\cmark & \cmark  & \textbf{85.0} & \textbf{87.0 }& \textbf{91.9} \\
			\bottomrule
		\end{NiceTabular}
	}
\end{table} 	
	
\subsection{Correct Candidate Helps Backward Reasoning
}
\label{sec:pred}

\begin{figure}[!t]
	\centering
	\!\!\!\!
	\subfigure[\label{fig:abl-comb-all-003}\textit{text-davinci-003}.]{\includegraphics[width=0.16\textwidth]{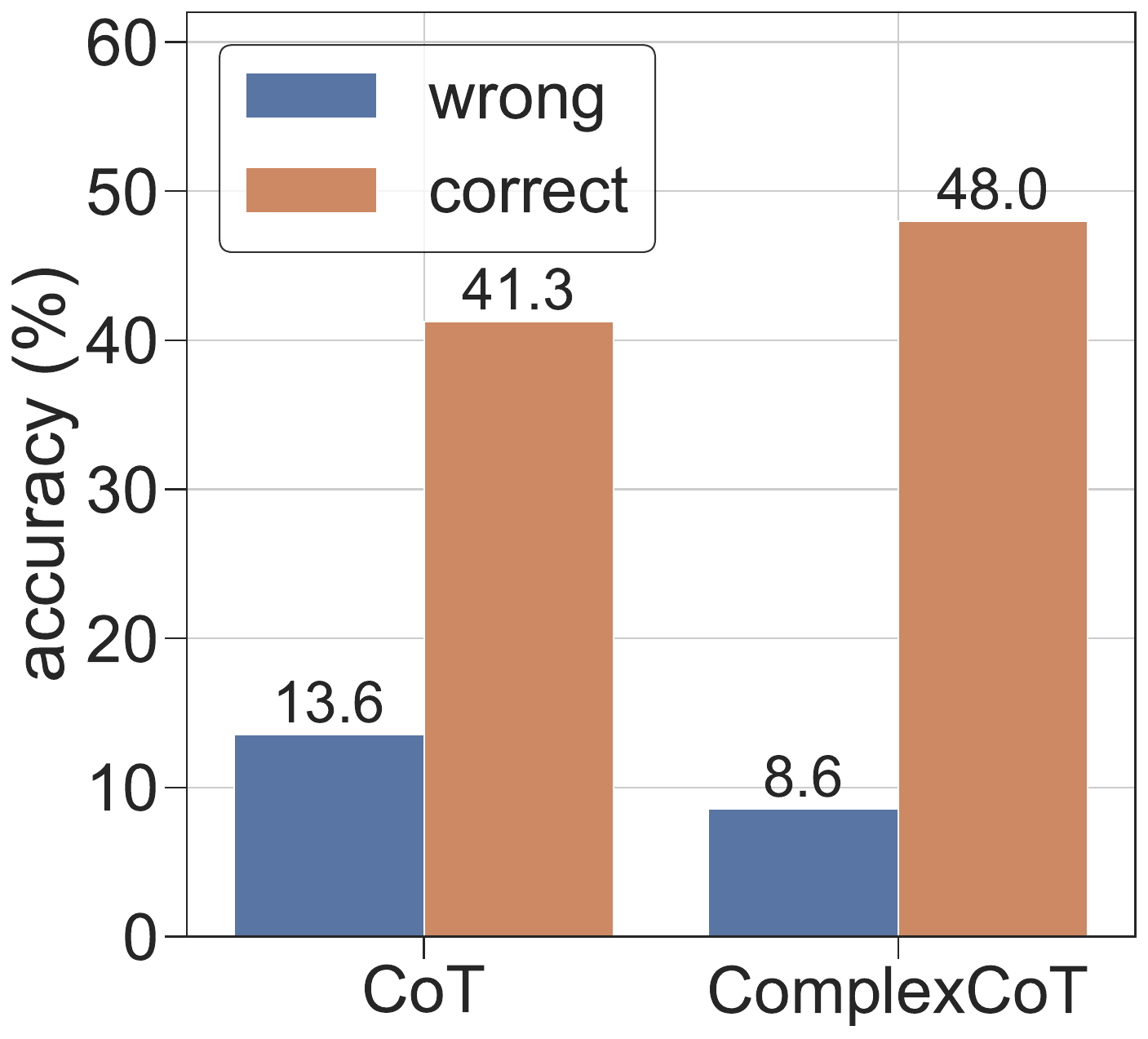}}
	\subfigure[\label{fig:abl-comb-all-turbo}\textit{GPT-3.5-Turbo}.]{\includegraphics[width=0.16\textwidth]{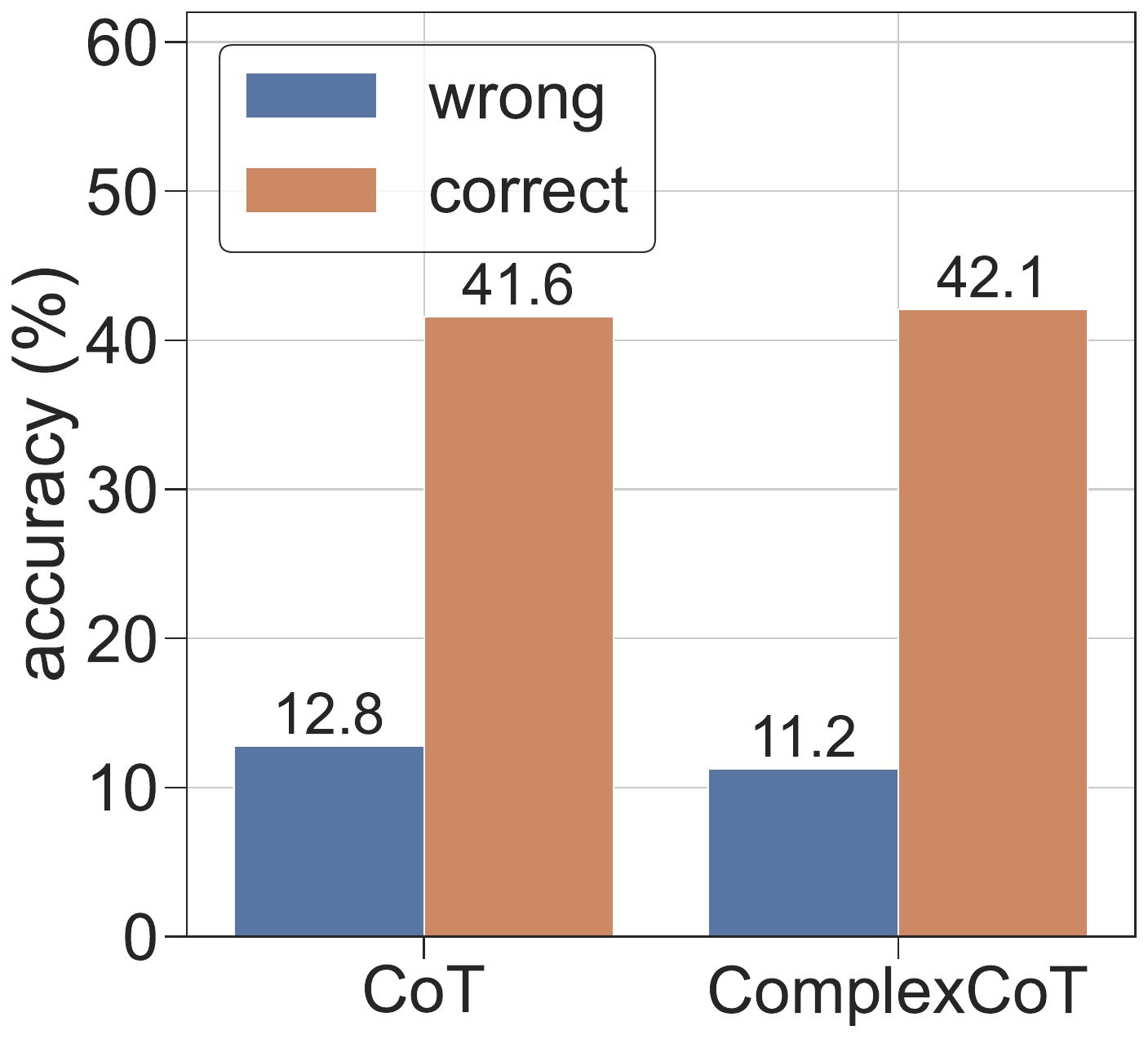}}
	\!
	\subfigure[\label{fig:abl-comb-all-gpt4}\textit{GPT-4}.]{\includegraphics[width=0.16\textwidth]{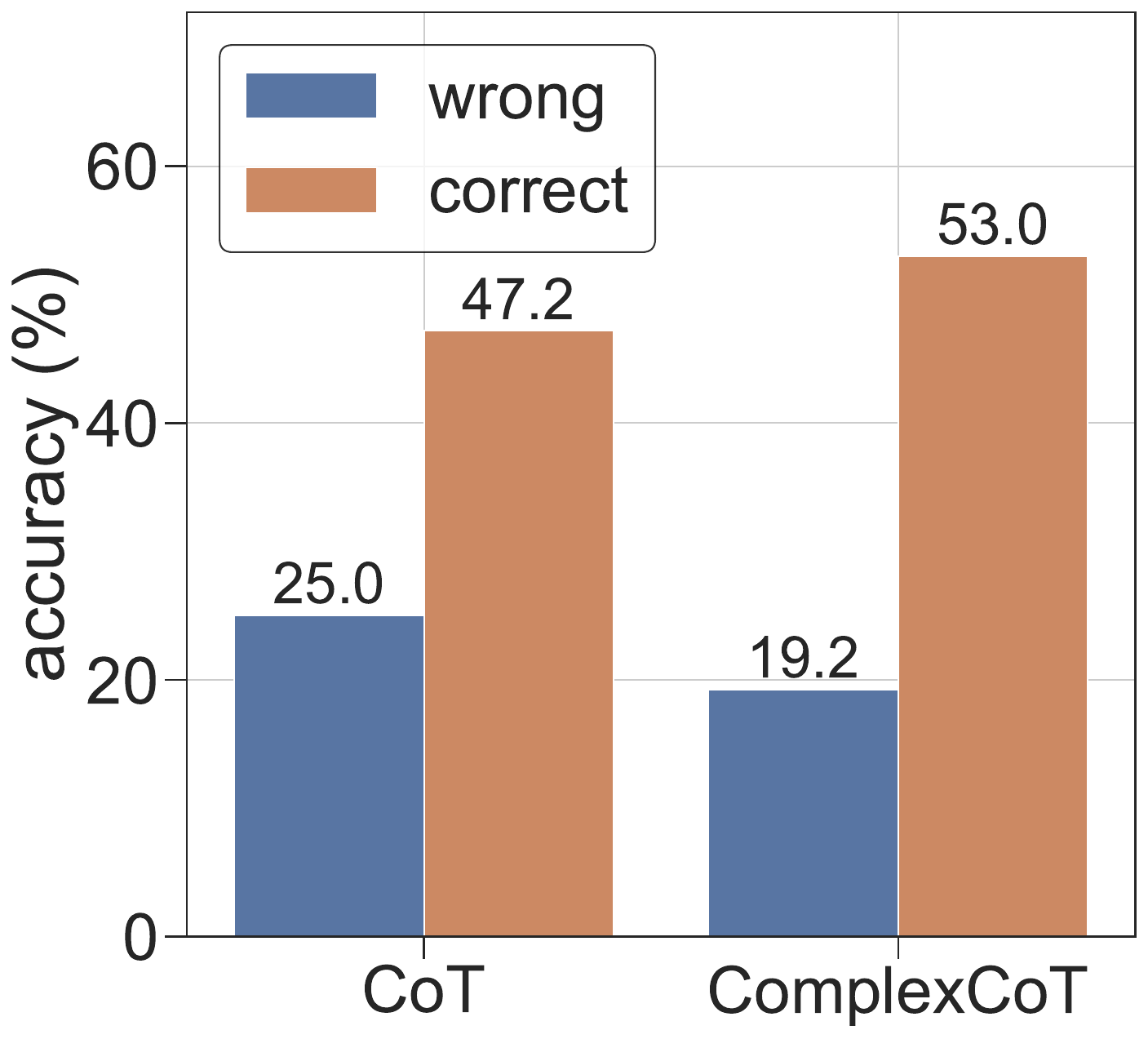}}
	\!\!\!\!
	\vskip -.2in
	\caption{Accuracy (averaged
		over all backward questions across the six data sets) of predicting the masked number in backward questions with correct/wrong candidate answers.}
	\label{fig:backward-acc}
\end{figure}

In this experiment, we verify the intuition that the correct candidate answer helps LLM to predict the masked numbers.
Figure \ref{fig:backward-acc} compares the accuracies of predicting the masked numbers in backward questions with the correct/wrong candidates.
As can be seen, using the correct candidate has $2\times$ higher accuracy (averaged over the six data sets) than the wrong ones in predicting masked numbers,
demonstrating that using backward reasoning to verify candidate answers is reasonable.

\subsection{Number of  Forward and Backward Reasoning Chains}
\subsubsection{Varying $M_\text{F}$}
\label{subsec:effect-Mf}

\begin{figure}[!t]
    \centering
    \vskip -.1in
    \!\!\!\!
    \subfigure[\label{fig:abl-Mf-all-complexcot-003}\textit{text-davinci-003}.]{\includegraphics[width=0.16\textwidth]{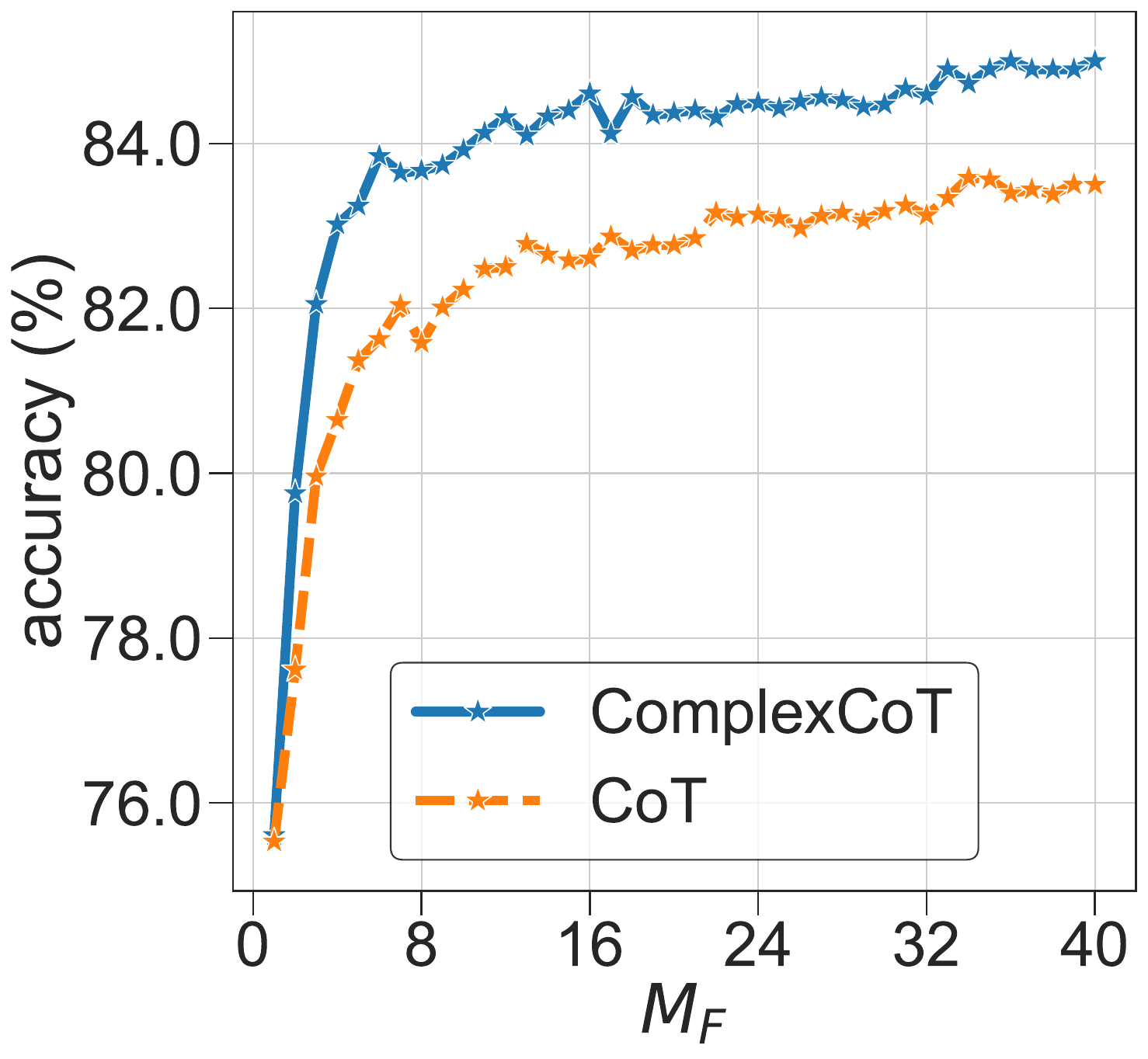}} \!
    \subfigure[\label{fig:abl-Mf-all-complexcot-3.5turbo}\!\! \textit{GPT-3.5-turbo}.]{\includegraphics[width=0.16\textwidth]{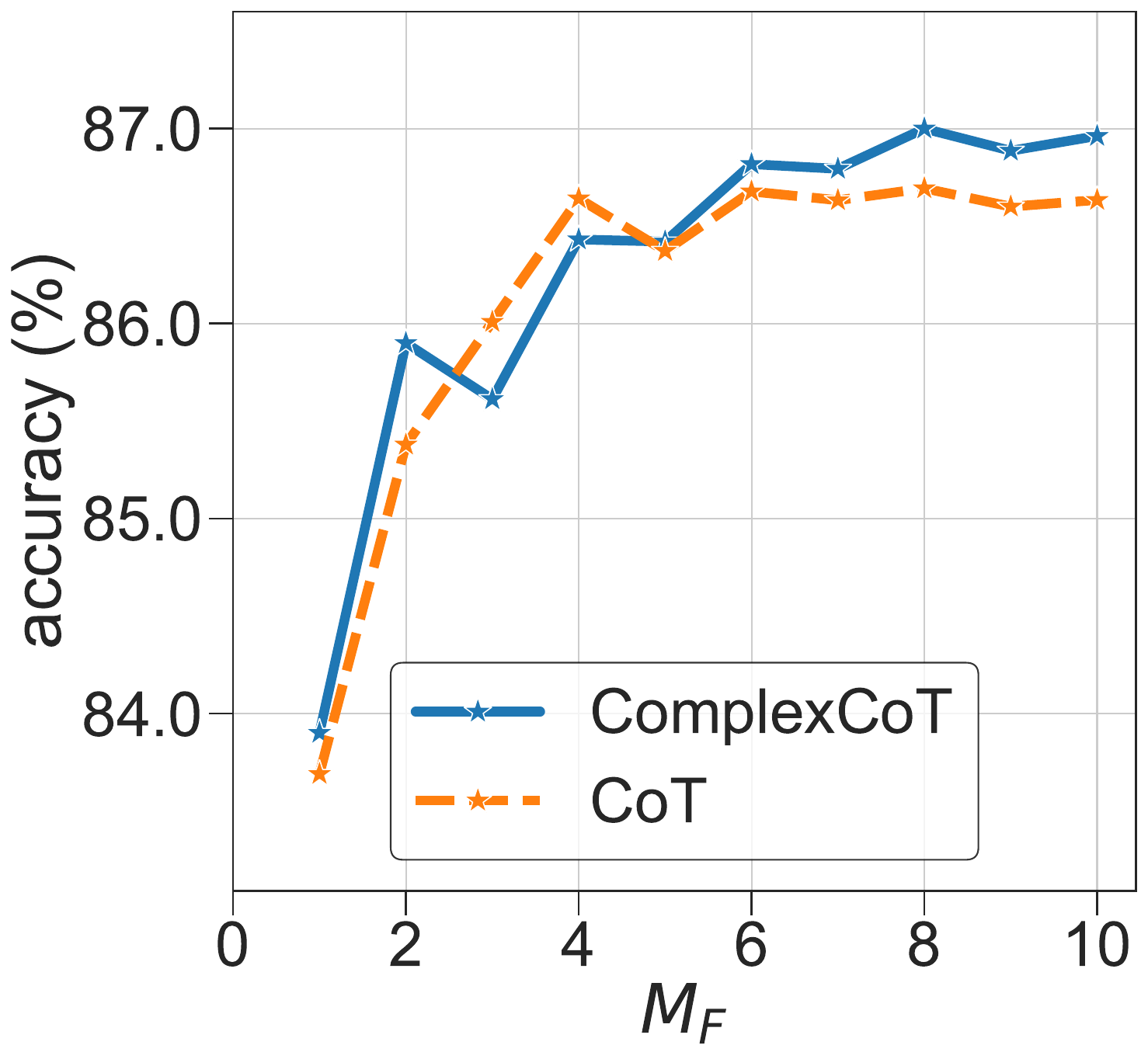}} \!
    \subfigure[\label{fig:abl-Mf-all-complexcot-gpt4}\!\! \textit{GPT-4}.]{\includegraphics[width=0.16\textwidth]{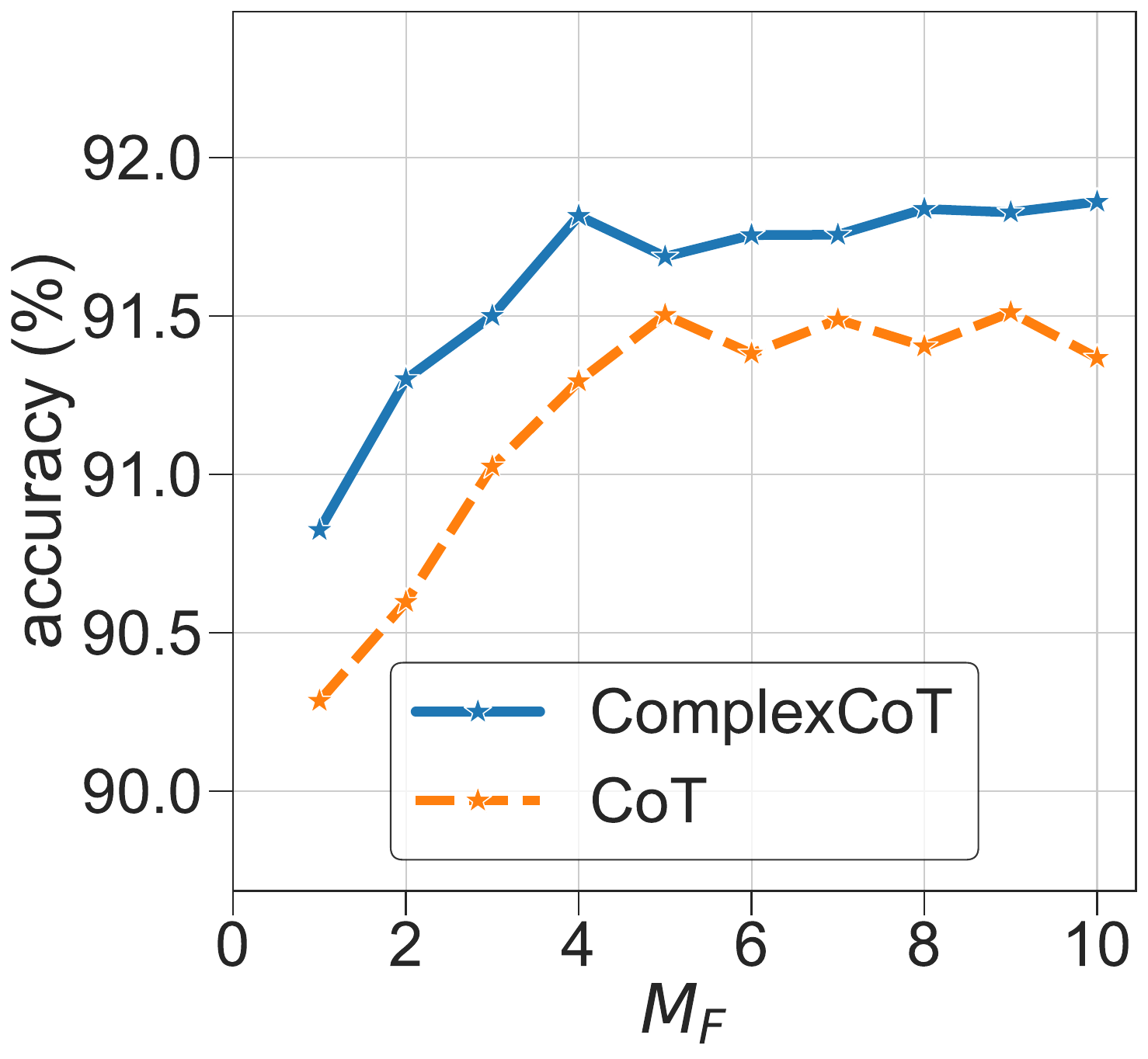}} \!\!\!\!
    \vskip -.18in
    \caption{
        Testing accuracy of FOBAR
        (averaged
        over the six data sets)
        with $M_\text{F}$.}
    \label{fig:albation-MF-all}
    \vskip -.2in
\end{figure}

\begin{figure}[!t]
	\centering
	\!\!\!\!
	\subfigure[\!\label{fig:all-sc-003}\textit{text-davinci-003}.]{\includegraphics[width=0.16\textwidth]{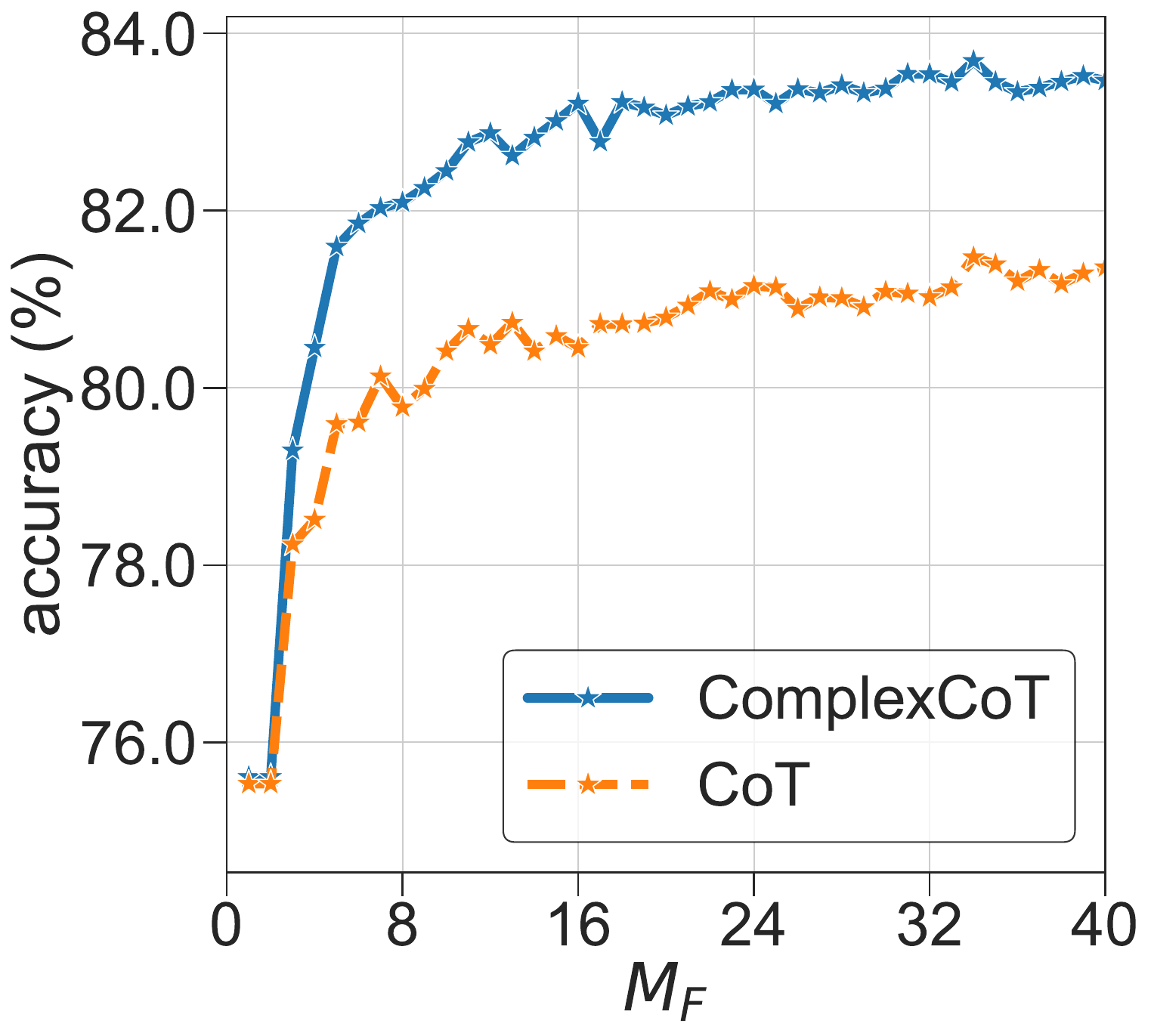}}
	\subfigure[\label{fig:all-sc-3.5turbo}\!\! \textit{GPT-3.5-Turbo}.]{\includegraphics[width=0.16\textwidth]{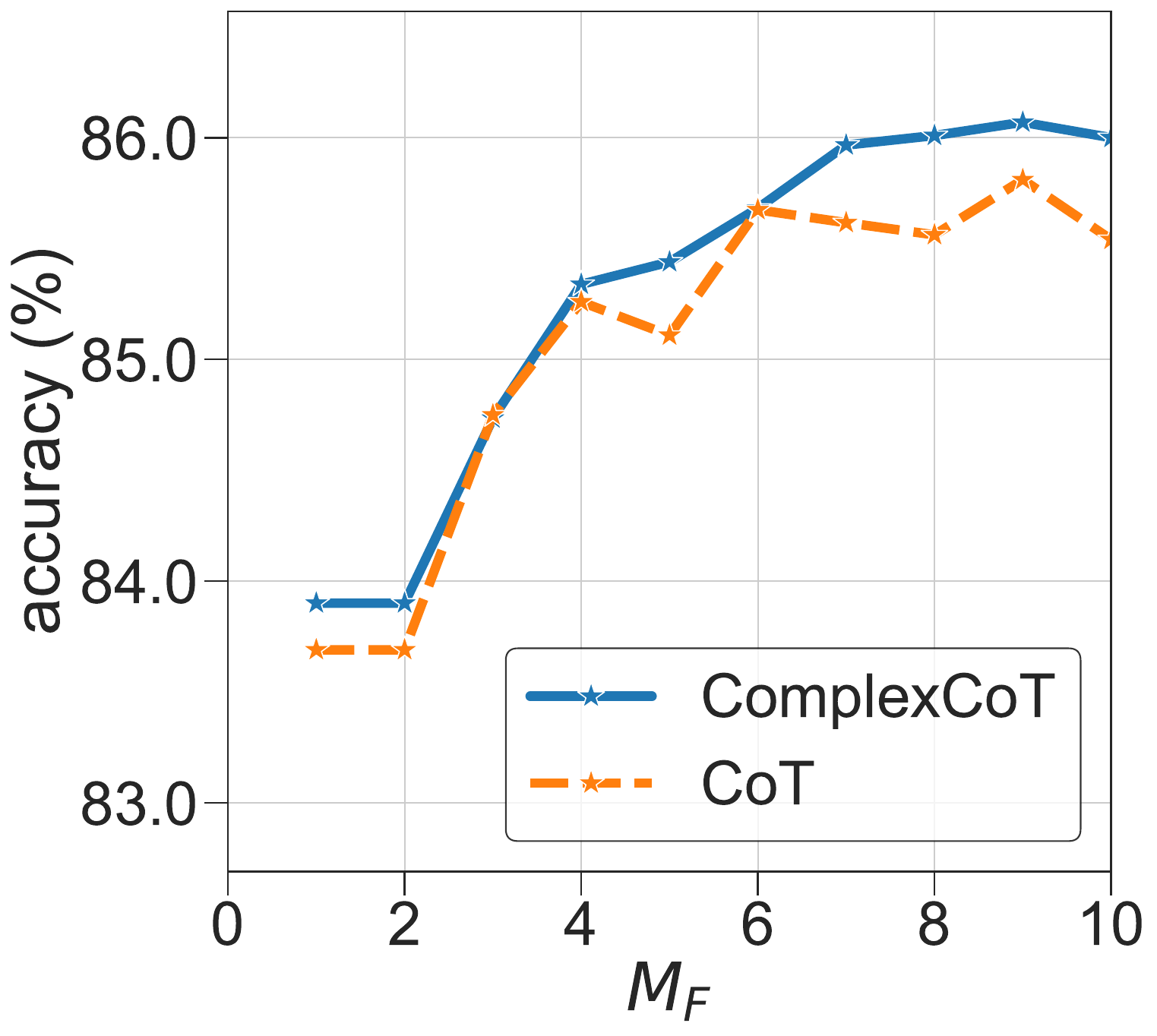}}\!
	\subfigure[\label{fig:all-sc-gpt4} \textit{GPT-4}.]{\includegraphics[width=0.16\textwidth]{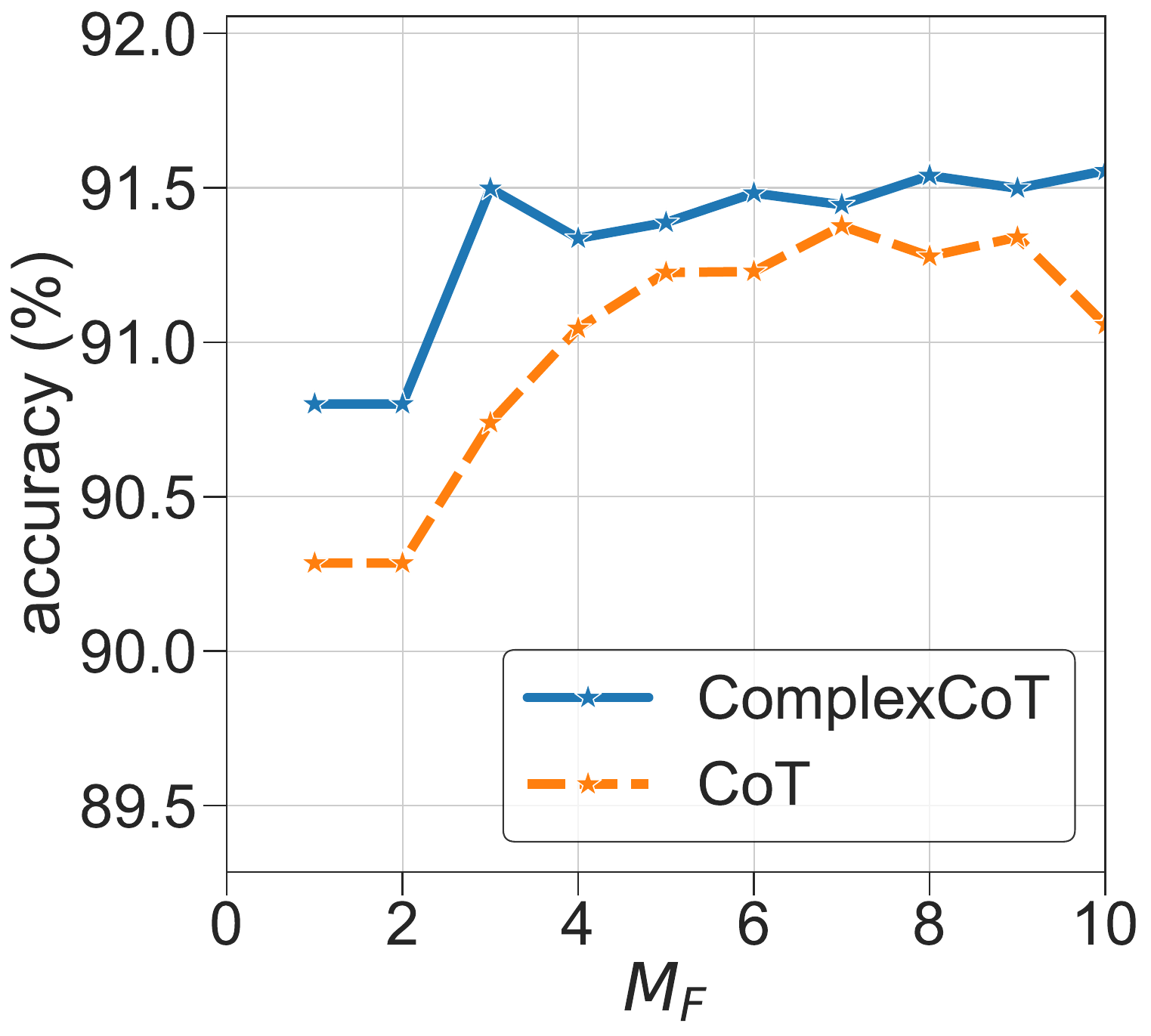}}\!\!\!\!
	\vskip -.2in
	\caption{Testing accuracy (averaged over six data sets) of Self-Consistency versus number of sampling
		paths ($M_\text{F}$).}
	\label{fig:sc-turbo-all}
	\vskip -.2in
\end{figure}

In this section, we study how the performance of
FOBAR
varies with the number of forward reasoning chains
$M_\text{F}$.
Figure~\ref{fig:albation-MF-all}
shows the
testing accuracies
(averaged over the six data sets)
for the three LLMs.
As can be seen, using a very small
$M_\text{F}$
(e.g., $\leq 5$) is clearly undesirable, but the accuracy
saturates quickly with increasing $M_\text{F}$.
This suggests that one
can use a small $M_\text{F}$
to reduce the computational cost.
Moreover,  
the accuracy curves of FOBAR are higher than
those of Self-Consistency in Figure~\ref{fig:sc-turbo-all},
again demonstrating that integrating backward reasoning into verification is effective.
	
\subsubsection{Varying $M_\text{B}$}
\label{subsec:effect-Mb}
\begin{figure}[!t]
    \centering
    \!\!\!\!
    \subfigure[\label{fig:abl-Mb-all-complexcot-003}\textit{text-davinci-003}.]{\includegraphics[width=0.16\textwidth]{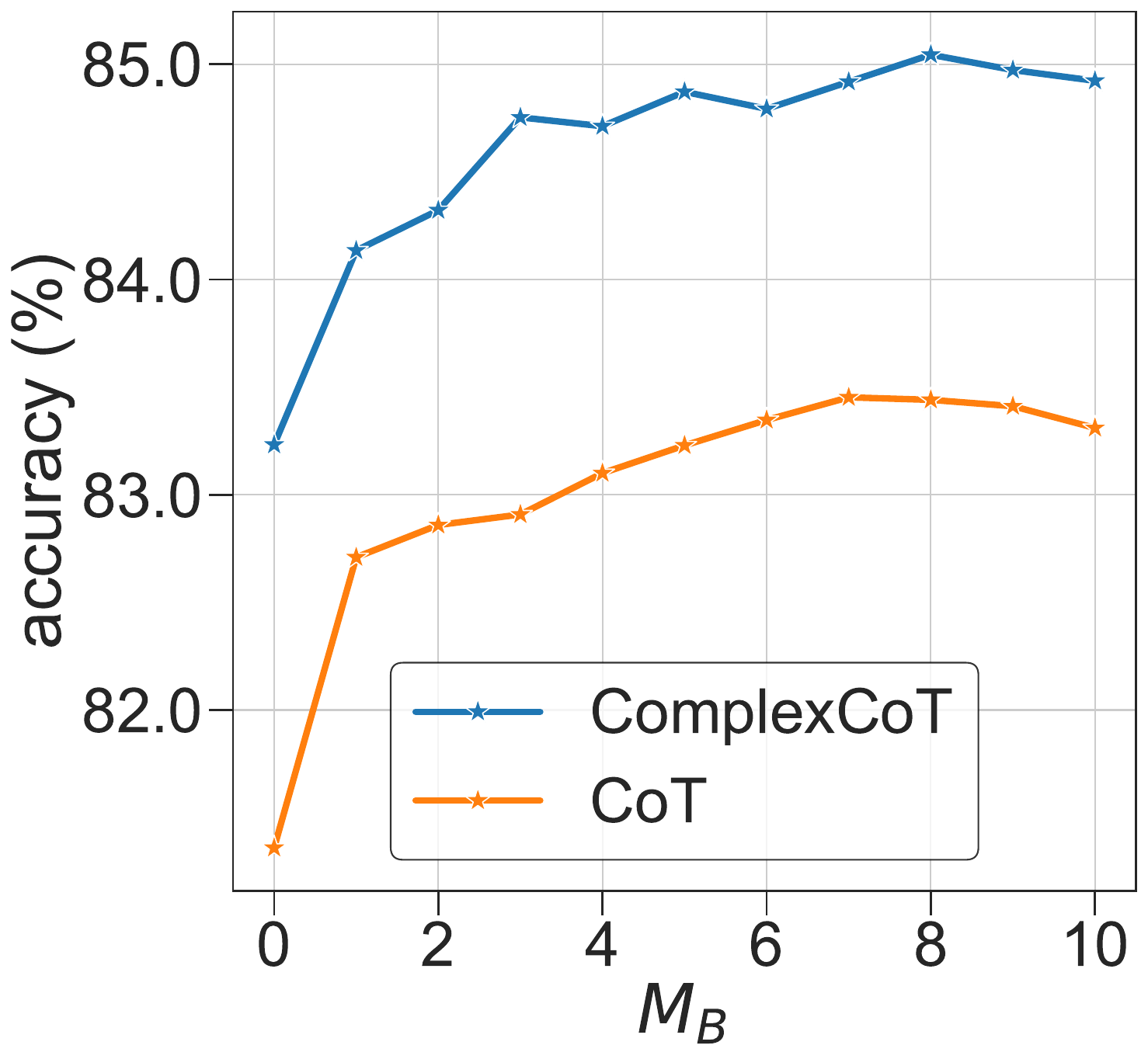}} 
    \subfigure[\label{fig:abl-Mb-all-complexcot-3.5turbo}\!\! \textit{GPT-3.5-turbo}.]{\includegraphics[width=0.16\textwidth]{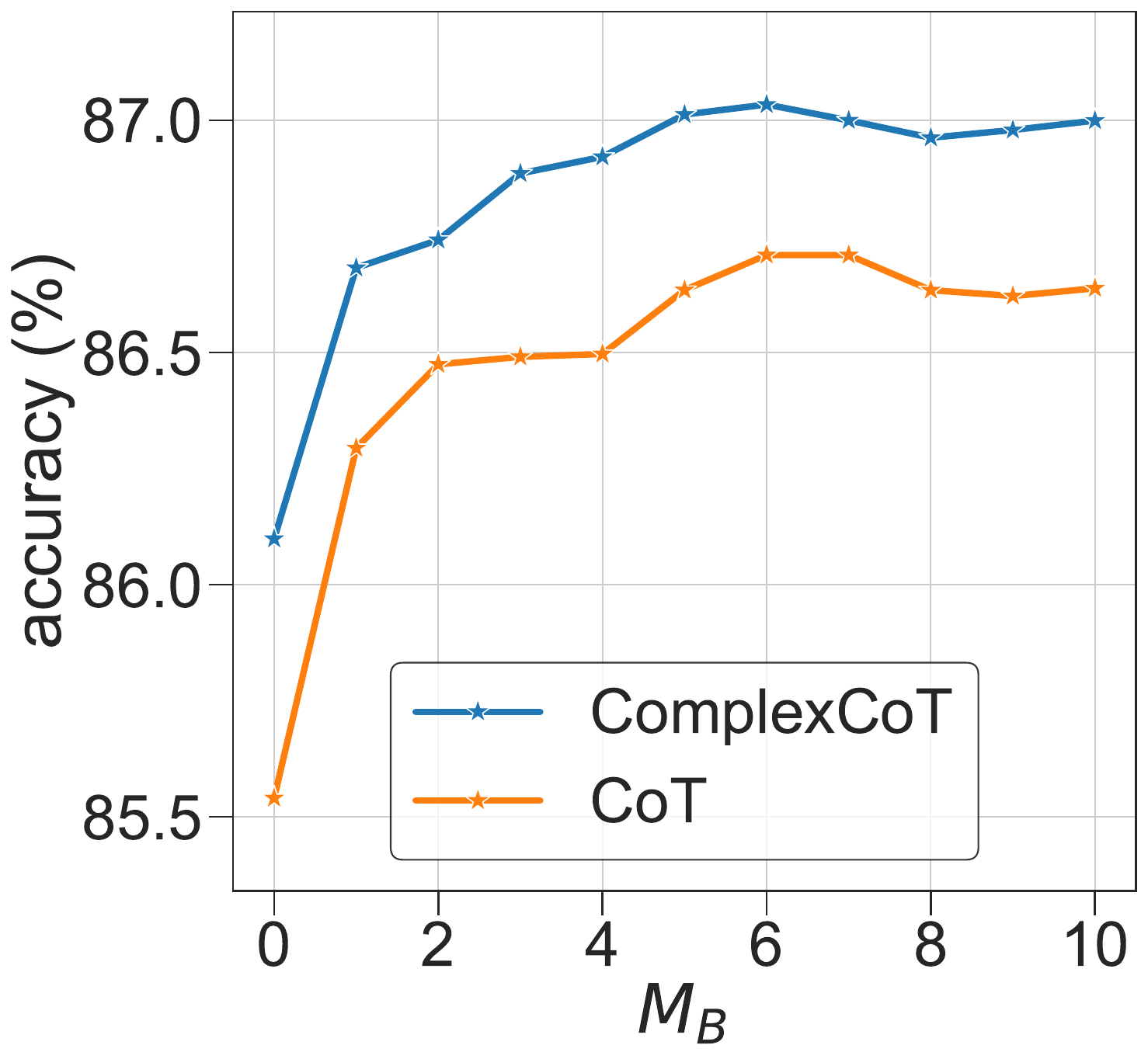}} \!
    \subfigure[\label{fig:abl-Mb-all-complexcot-gpt4}\!\! \textit{GPT-4}.]{\includegraphics[width=0.16\textwidth]{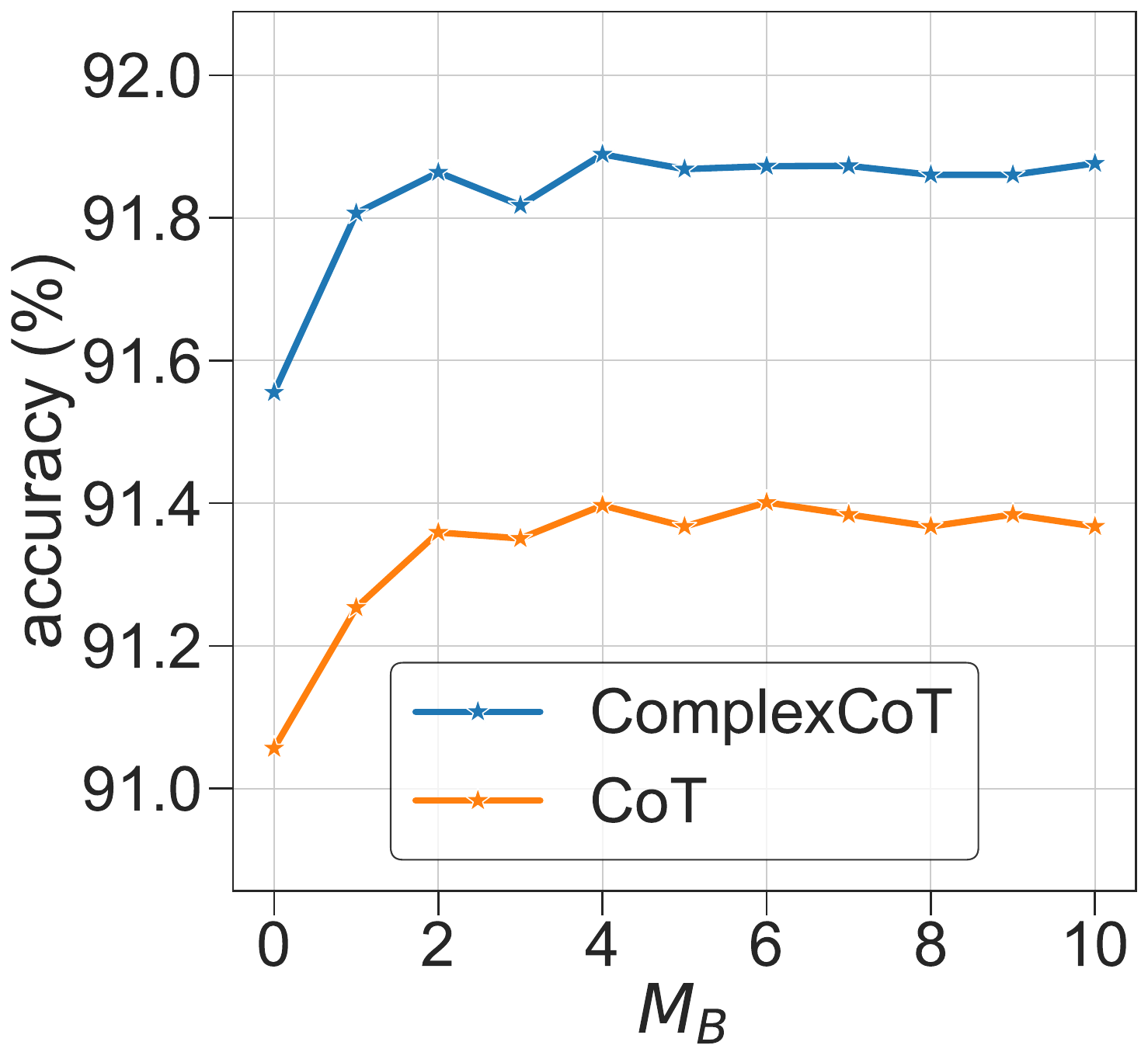}} \!\!\!\!
    \vskip -.18in
    \caption{
        Testing accuracy of FOBAR
        (averaged
        over the six data sets)
        with $M_\text{B}$.}
    \label{fig:albation-Mb-all}
\end{figure}

Next,
we study how the performance of
FOBAR varies with the number of 
backward 
reasoning chains
$M_\text{B}$.
Figure~\ref{fig:albation-Mb-all}
shows the
testing accuracies
(averaged over the six data sets)
for the three LLMs.
Note that $M_\text{B}=0$ corresponds to using
only
forward reasoning.
As shown, using a very small
$M_\text{B}$
(e.g., $\leq 4$)
is clearly undesirable, but the accuracy saturates quickly when $M_\text{B}$ increases.
Hence,
using a small $M_\text{B}$ can achieve a good balance between performance and
efficiency.

\begin{table}[!t]
	\centering
	\vskip -.05in
	\caption{Accuracies on the non-mathematical tasks of
		\textit{Date Understanding}
		(denoted \textit{DateU}) and \textit{Last Letter Concatenation}
		(denoted \textit{LastLetter})
		using \textit{GPT-3.5-Turbo}.
		Results with $^\dagger$ are from the original publications.
		``--'' means that the result is not reported in the original publication.
	}
	\vskip -.15in
	\renewcommand{\arraystretch}{1.1}
	\label{table:date-letter}
	\resizebox{.48\textwidth}{!}{
		\begin{NiceTabular}{cccc}
			\CodeBefore
			\rectanglecolor{orange3!50}{3-1}{7-1}
			\rectanglecolor{blue3!50}{8-1}{12-1}
			\rectanglecolor{Gray}{7-2}{7-4}
			\rectanglecolor{Gray}{12-2}{12-4}
			\Body
			\toprule
			& & \textit{DateU} & \textit{LastLetter}\\
			\midrule
			& ICL~\citep{brown2020language} & 52.0& 8.0\\
			\midrule
			\multirow{5}{*}{\STAB{\rotatebox[origin=c]{90}{CoT}}}  & 
			CoT~\citep{wei2022chain} & 61.3 & 81.0  \\
			& RE2$^\dagger$~\citep{xu2023re} & 47.2 & - \\
			& Self-Consistency~\citep{wang2023selfconsistency} & 65.6 & 81.4 \\
			& Self-Verification~\citep{weng2022large} &66.1& 81.8\\
			& FOBAR & \textbf{66.4} & \textbf{82.6}\\
			\midrule
			\multirow{5}{*}{\STAB{\rotatebox[origin=c]{90}{ComplexCoT}}}  
			& ComplexCoT~\citep{fu2023complexitybased} & 74.8 & 81.4 \\
			& RCoT$^\dagger$~\citep{xue2023rcot} & 71.7 & - \\
			& Self-Consistency~\citep{wang2023selfconsistency} & 77.5 & 81.2 \\
			& Self-Verification~\citep{weng2022large} & 76.2 & 81.6 \\
			& FOBAR & \textbf{78.0} & \textbf{82.4} \\
			\bottomrule
		\end{NiceTabular}
	}
	\vskip -.2in
\end{table} 

\begin{table*}[!h]
	\centering
	\caption{Statistics on the failure cases of Self-Consistency on the six data sets.}
	\label{tbl:sc-fail-analysis}
	\vskip -.15in
	\resizebox{.9\textwidth}{!}{
		\begin{tabular}{cccccccc}
			\toprule
			& \textit{AddSub} & \textit{MultiArith} & \textit{SingleEQ} & \textit{SVAMP} & \textit{GSM8K} & \textit{AQuA} & Total\\
			\midrule
			\#failures & 47 & 7 & 28 & 150 & 179 & 94 & 505\\
			\midrule
			\#failures with no correct answer& 28 & 0 & 14 & 57 & 60 & 52 & 211 \\
			\#failures with at least one correct answer& 19 & 7 & 14 & 93 & 119 & 42 & 294 \\
			\bottomrule
		\end{tabular}
	}
\end{table*}

\begin{table*}[!h]
	\centering
	\caption{Statistics on the failure cases of FOBAR on the six data sets.}
	\label{tbl:fobar-fail-analysis}
	\vskip -.15in
	\resizebox{.9\textwidth}{!}{
		\begin{tabular}{cccccccc}
			\toprule
			& \textit{AddSub} & \textit{MultiArith} & \textit{SingleEQ} & \textit{SVAMP} & \textit{GSM8K} & \textit{AQuA} & Total\\
			\midrule
			\#failures & 46 &  1 & 29 & 115 & 166 &  94& 451 \\
			\midrule
			\#\#failures with no correct answer & 28 & 0 & 14 & 57 & 60 & 52 & 211 \\
			\#failures with at least one correct answer& 18 & 1 & 15 & 58 & 106 & 42 & 240 \\
			\bottomrule
		\end{tabular}
	}
\end{table*}

\subsection{Extension to Non-Mathematical Tasks}
\label{sec:non-math}

In this section,
we perform experiments on two commonly-used
non-mathematical tasks: \textit{Date Understanding}~\citep{wei2022chain, fu2023complexitybased} 
and \textit{Last Letter Concatenation}~\citep{wei2022chain, zhou2023leasttomost}.
Examples are shown in Table~\ref{table:data sets} (Appendix \ref{appendix:dataset}).
We compare FOBAR with other CoT-based methods and ICL using \textit{GPT-3.5-Turbo}. PHP does not report results on non-mathematical tasks. 

Table~\ref{table:date-letter}
shows the testing accuracies.
As shown,
FOBAR performs better than all the baselines with either CoT or ComplexCoT as base prompt.
Moreover, all
CoT-based methods 
significantly outperform ICL.

\subsection{Failure Cases of Self-Consistency and FOBAR}
\label{apd:sc}

We conduct an analysis on the failure cases of Self-Consistency and FOBAR on the six data sets, using \textit{GPT-3.5-Turbo} with ComplexCoT prompting.
Table~\ref{tbl:sc-fail-analysis}
shows the number of failure cases of Self-Consistency, with a breakdown into 
the numbers of cases with no chain reaching the correct answer
and at least one chain reaching the correct answer.
As can be seen, 
about 60\% of the total failure cases have at least one correct chains
(the remaining 40\% have no correct chains and thus cannot be solved by backward reasoning).
These
60\% cases can potentially be fixed with a better verifìer (such as the proposed FOBAR).
Table~\ref{tbl:fobar-fail-analysis}
shows the statistics on the failure cases of FOBAR.
As can be seen,
FOBAR rectifies 
54 (i.e., $294-240$) out of the 294 failure cases that have at least one correct answer in Self-Consistency.

\section{Conclusion}
\label{sec:conclusion}
	
In this paper,
we study the problem of verifying candidate answers to 
mathematical problems using chain-of-thought prompting.
To complement the use of only forward reasoning
for verification,
we introduce
backward reasoning:
A simple template is introduced to create questions and a prompt is
designed to ask the LLM to predict a masked word when a candidate answer is provided.
Furthermore, we
proposed FOBAR
to combine forward and backward reasoning for verification.
Extensive experiments 
on six standard mathematical data sets and
three LLMs
show that the proposed FOBAR
achieves state-of-the-art performance on mathematical reasoning tasks.
FOBAR 
can also be used
on non-mathematical tasks and achieves superior performance.
	
\section*{Limitations and Potential Risks}

\paragraph{Limitations}
In this paper, we focused on mathematical reasoning tasks, with extension to two non-mathematical reasoning tasks.
However, extensions to more complicated non-mathematical reasoning tasks such as Common-Sense Question-Answering (CSQA)~\citep{wei2022chain}
and StrategyQA~\citep{wei2022chain,fu2023complexitybased} are still to be explored, as 
identifying the informative words to mask is more challenging.
 
When a number is superfluous in the question (unnecessary in solving the question), the number is probably unpredictable when a candidate answer is provided. Hence, the superfluous numbers may not affect the number of correct backward chains $Z_c$'s, which mainly depend on the critical numbers. Thus, FOBAR is still applicable. Though it is more accurate to avoid masking redundant numbers, checking whether a number is redundant is challenging and will be studied in our future work.

\paragraph{Potential Risks}
All data sets used in this work 
do not contain any information that names or uniquely identifies individual people or offensive content.
Hence,
there is no
concern about ethical considerations and data privacy.

\section*{Acknowledgement}
This work was supported by NSFC key grant 62136005, NSFC general grant 62076118, and Shenzhen
fundamental research program JCYJ20210324105000003. This research was supported in part by the
Research Grants Council of the Hong Kong Special Administrative Region (Grants 16200021 and
16202523).
	
{\small
\bibliography{paper_cr}
}

\clearpage
\appendix

\section{Question-Answer Demos of Backward Reasoning}
\label{sec:exmp-back}

Example \ref{exmp:inv-q}
shows three backward questions that mask different numbers in the original question.
Example \ref{exmp:backward} shows a  backward question and its answer.

    \begin{exmp}{Backward questions.}{inv-q}
    \textbf{Question}:
    Jim spends {\color{red3}\normalfont x} hours watching TV and then decides to go to bed and reads for half as long.  He does this 3 times a week.  How many hours does he spend on TV and reading in 4 weeks?
    \textit{If we know the answer to the above question is $\{\hat{A}_c\}$, what is the value of unknown variable {\normalfont x}?}
    
    \textbf{Question}:
    Jim spends 2 hours watching TV and then decides to go to bed and reads for half as long.  He does this {\color{red3}x} times a week.  How many hours does he spend on TV and reading in 4 weeks? \textit{If we know the answer to the above question is $\{\hat{A}_c\}$, what is the value of unknown variable {\normalfont x}?}
    
    \textbf{Question}:
    Jim spends 2 hours watching TV and then decides to go to bed and reads for half as long.  He does this 3 times a week.  How many hours does he spend on TV and reading in {\color{red3}\normalfont x} weeks?
    \textit{If we know the answer to the above question is $\{\hat{A}_c\}$, what is the value of unknown variable {\normalfont x}?}
\end{exmp}

\begin{exmp}{Backward Reasoning.}{backward}
	\textbf{Question:} Randy has 60 mango trees on his farm. He also has x less than half as many coconut trees as mango trees. How many trees does Randy have in all on his farm?
	\textit{If we know the answer to the above question is 85, what is the value of unknown variable {\normalfont x}?}
	
	\textbf{Answer:}
	Let’s think step by step.
	We know that Randy has 60 mango trees on his farm.
	We also know that he has x less than half as many coconut trees as mango trees. Let's use C to represent the number of coconut trees.
	So we can write: C = (1/2)*60 - x = 30 - x.
	The total number of trees on Randy's farm is the sum of the number of mango trees and coconut trees: 60 + (30 - x) = 90 - x.
	We are given that the total number of trees on Randy's farm is 85, so we can write: 90 - x = 85.
	Solving for x, we get: x = 5.
	The value of x is 5.
\end{exmp}

\section{Example Rewriting Mistake in Self-Verification}
\label{appendix:fail-sv}

\begin{emptybox}{}{}
    \!\!\!\!\!\!
    \textbf{Question:}
    A class of 50 students has various hobbies. 10 like to bake, 5 like to play basketball, and the rest like to either play video games or play music. How many like to play video games if the number that like to play music is twice the number that prefer playing basketball? {\color{green3}(answer: 25)}
\end{emptybox}

We mask the first number (i.e., 50) by $\vx$ and a candidate answer 25 is provided.
The following shows the backward questions obtained
by Self-Verification
and FOBAR.
We can see that Self-Verification
makes a mistake in rewriting the question into a declarative statement,
while the proposed simple template in FOBAR does not need rewriting.

\begin{emptybox}{}{}
    \!\!\!\!\!\!
    \textbf{Question (Self-Verification):}
    A class of x students has various hobbies. 10 like to bake, 5 like to play basketball, and the rest like to either play video games or play music. {\color{red3}The number of people who like to play video games is equal to the number of people who prefer playing basketball multiplied by two.} The number of people who like to play video games is 25.
    What is the answer of x?
    
    \textbf{Question (FOBAR):}
    A class of x students has various hobbies. 10 like to bake, 5 like to play basketball, and the rest like to either play video games or play music. How many like to play video games if the number that like to play music is twice the number that prefer playing basketball?
    \textit{If we know the answer to the above question is 25, what is the value of unknown variable {\normalfont x}?}
\end{emptybox}

\section{Example Cases showing that Forward and Backward Reasoning are Complementary}
\label{apd:example-bw-fw}

In this section,
we show that forward and backward reasoning are complementary,
i.e.,
failure cases in forward reasoning can be corrected by backward reasoning, and vice versa.
We use cases from the \textit{SingleEQ} data set using
\textit{text-davinci-003} with CoT prompting.
Example~\ref{exmp:fw-w-bw-r}
shows a case where forward reasoning (i.e., Self-Consistency) fails but backward reasoning succeeds.
We can see that this problem is difficult to solve in the forward direction,
but the correctness of a candidate answer can be easily verified in the backward direction.
Example~\ref{exmp:fw-r-bw-w}
shows a case where backward reasoning fails 
but forward reasoning succeeds.
Moreover,
FOBAR can choose the correct answer in both cases.

\begin{exmp}{Forward reasoning fails but backward reasoning succeeds.}{fw-w-bw-r}
\textbf{Question:}
The sum of three consecutive odd numbers is 69. What is the smallest of the three numbers?

\noindent
\textbf{Ground-truth answer}: 21

\textbf{Forward reasoning:} 
$\bP_\text{F}(21)=0.4, \bP_\text{F}(23)=0.6$

\textbf{Backward reasoning:}
$\bP_\text{B}(21)=0.8, \bP_\text{B}(23)=0.2$

\textbf{FOBAR:}
$\bP(21)=0.62, \bP(23)=0.38$

\textbf{A backward question:}
The sum of three consecutive odd numbers is ${\bf x}$. What is the smallest of the three numbers?
If we know the answer to the above question is $21$, what is the value of unknown variable ${\bf x}$?
\end{exmp}

\begin{exmp}{Forward reasoning succeeds but backward reasoning fails.}{fw-r-bw-w}
	\textbf{Question:} While digging through her clothes for ice cream money, Joan found 15 dimes in her jacket, and 4 dimes in her shorts. How much money did Joan find?
	
	\textbf{Ground-Truth answer:} 1.9
	
	\textbf{Forward reasoning:}
	$\bP_\text{F}(1.9)=0.7, \bP_\text{F}(190)=0.3$
	
	\textbf{Backward reasoning}: 
	$\bP_\text{B}(1.9)=0.43, \bP_\text{B}(190)=0.57$
	
	\textbf{FOBAR:}
	$\bP(1.9)=0.57, \bP(190)=0.43$

\textbf{A backward question:}
While digging through her clothes for ice cream money, Joan found 15 dimes in her jacket, and $\vx$ dimes in her shorts. How much money did Joan find?
If we know the answer to the above question is $1.9$, what is the value of unknown variable ${\bf x}$?
\end{exmp}

\section{Additional Experiments}

\subsection{Comparison between FOBAR and Trained Verifiers}
\label{sec:train-verfier}

Compared with \citet{cobbe2021training}, which trains an LLM for verifying answers, FOBAR has two advantages. 
(i) \textbf{(training-free)} Training an LLM for verification is computationally expensive and labor-intensive in collecting extra annotation data, while backward reasoning for verification is training-free and requires no additional data collection. 
(ii) \textbf{(more effective)} As training the GPT-3 (175B) model is extremely expensive and their code is not publicly available, we compare our FOBAR with the result reported in Figure 5 of \citep{cobbe2021training}, where the candidate answers are generated by \textit{GPT-3}. Table \ref{table:baseline} shows the accuracy of \textit{GSM8K}. As shown, \mbox{FOBAR} consistently performs much better than the trained verifier (\textbf{+14.8}).  

\begin{table}[!t]
   \centering
   \caption{Comparison between \mbox{FOBAR} and a trained verifier on \textit{GSM8K}.}
   \label{table:baseline} 
   \vskip -.15in
   \resizebox{.48\textwidth}{!}{
   \begin{tabular}{lc}
       \toprule
       Training GPT-3 (175B) for Verification \citep{cobbe2021training} & 56.0  \\ \rowcolor{Gray}
       FOBAR (text-davinci-003 + CoT) & 70.8 \\
       \rowcolor{Gray}
       FOBAR (text-davinci-003 + ComplexCoT) & 78.7 \\ \rowcolor{Gray}
       FOBAR (GPT-3.5-Turbo + CoT) & 85.1 \\ \rowcolor{Gray}
       FOBAR (GPT-3.5-Turbo + ComplexCoT) & 87.4 \\
       \rowcolor{Gray}
       FOBAR (GPT-4 + CoT) & 95.4 \\ \rowcolor{Gray}
       FOBAR (GPT-4 + ComplexCoT) & 96.4 \\
       \bottomrule
   \end{tabular}
}
\end{table}

\subsection{Comparison between FOBAR and Step-by-Step Forward Verification}
\label{apd:step-step-forward}

Recent works \citep{lightman2023let, ling2023deductive}
propose verifying the steps of forward reasoning chains.
\citet{lightman2023let}
propose to label exclusively steps of forward reasoning chains generated by LLMs.
The labeled data are then used to train an LLM
for verification.
Compared with \citep{lightman2023let}, which is computationally expensive in training an LLM and labor-intensive in labeling data,
our backward 
reasoning is training-free for verification and 
requires no additional data annotation. 

\begin{table}[!t]
	\centering
	\caption{Accuracy of FOBAR when combining backward reasoning with three types of forward reasoning for verification. BR stands for ``Backward Reasoning''.}
	\label{table:ded-verx}
	\vskip -.15in
	\resizebox{.48\textwidth}{!}{
		\begin{tabular}{l c c c}
			\toprule
			& \textit{AddSub} & \textit{GSM8K} & \textit{AQuA} \\
			\midrule
			Self-Consistency  & 88.1 & 86.4 & 63.0 \\ \rowcolor{Gray}
			Self-Consistency + BR &\textbf{ 88.4} & \textbf{87.4} & \textbf{63.4}\\
			\midrule
			NP \citep{ling2023deductive} & 93.67 &  87.05 & 70.34 \\ \rowcolor{Gray}
			NP + BR & \textbf{93.92} & \textbf{87.89} & \textbf{71.65} \\
			\midrule
			NP + DV + UPV \citep{ling2023deductive} & 93.54 & 86.01 & 69.49  \\ \rowcolor{Gray}
			NP + DV + UPV + BR & \textbf{93.92} &  \textbf{87.19} & \textbf{70.86} \\				\bottomrule
		\end{tabular}
	}
\end{table}

\begin{table*}[!h]
	\centering
	\caption{Statistics of data sets used in the experiments.}
	\vskip -.15in
	\!\!\!
	\resizebox{.98\textwidth}{!}{
		\begin{NiceTabular}{c|cccc}
			\toprule
			&& \#samples & $N_Q$ (mean $\pm$ std)& example  \\
			\midrule
			\multirow{15}{*}{\STAB{\rotatebox[origin=c]{90}{Math}}}  
			&\textit{AddSub} & 395 & $ 2.6 \pm 0.7$ & \demobox{ Benny picked 2 apples and Dan picked 9 apples from the apple tree. How many apples were picked in total?}  \\
			\cmidrule{2-5}
			&\textit{MultiArith} & 600  & $ 3.1 \pm 0.3 $ & \demobox{ Katie picked 3 tulips and 9 roses to make flower bouquets. If she only used 10 of the flowers though, how many extra flowers did Katie pick?} \\
			\cmidrule{2-5}
			&\textit{SingleEQ}  & 508 & $ 2.2 \pm 0.7 $ & \demobox{ Joan went to 4 football games this year. She went to 9 football games last year. How many football games did Joan go to in all?} \\
			\cmidrule{2-5}
			&\textit{SVAMP} &  1000 & $ 2.8 \pm 0.7 $ & \demobox{ Rachel has 4 apple trees. She picked 7 apples from each of her trees. Now the trees have a total 29 apples still on them. How many apples did Rachel pick in all?} \\
			\cmidrule{2-5}
			&\textit{GSM8K} & 1319  & $ 3.8 \pm 1.6 $ &\demobox{ A robe takes 2 bolts of blue fiber and half that much white fiber. How many bolts in total does it take?} \\
			\cmidrule{2-5}
			&\textit{AQuA} & 254 & $ 2.9 \pm 1.3 $ & \demobox{ If the population of a city increases by 5\% annually, what will be the population of the city in 2 years time if its current population is 78000? \newline
				Answer Choices: (A) 81900 (B) 85995 (C) 85800 (D) 90000 (E) None of these}  \\
			\midrule 
			\multirow{2}{*}{\STAB{\rotatebox[origin=c]{90}{Non-Math}}}  & 
			\textit{Last Letter Concatenation} & 500 & $4.0\pm 0.0$& \demobox{Take the last letters of each word in ``\textit{Whitney Erika Tj Benito}'' and concatenate them.} \\
			\cmidrule{2-5}
			&\textit{Date Understanding} & 369 & $1.2\pm 0.7$ & \demobox{The deadline is Jun 1, 2021, which is 2 days away from now. What is the date a month ago in MM/DD/YYYY?} \\
			\bottomrule
		\end{NiceTabular}
	}
	\label{table:data sets}
	\vskip -.15in
\end{table*}

\citet{ling2023deductive}
propose a natural language-based deductive reasoning format that allows the LLM to verify \textbf{forward} reasoning steps.
Different from \citep{ling2023deductive},
we use \textbf{backward} reasoning 
to verify the candidate answers instead of the steps in forward chains.
As
backward and forward reasoning are complementary,
the proposed
backward reasoning
can be combined with their step-by-step forward methods.
We replace the forward reasoning in FOBAR (i.e., Eq.~\eqref{eq:fobar}) with step-by-step verification proposed by \citet{ling2023deductive},
and conduct experiments on \textit{AddSub}, \textit{GSM8K}, and \textit{AQuA} using \textit{GPT-3.5-Turbo}. 
Table~\ref{table:ded-verx}
shows the testing accuracy.
As can be seen, combining backward reasoning with forward reasoning methods consistently boosts performance.

\section{Data Sets}
\label{appendix:dataset}

Table \ref{table:data sets}
shows the statistics on the data sets used in the experiments.
	
\end{document}